\documentclass[conference]{IEEEtran}
\IEEEoverridecommandlockouts

\def\BibTeX{{\rm B\kern-.05em{\sc i\kern-.025em b}\kern-.08em
    T\kern-.1667em\lower.7ex\hbox{E}\kern-.125emX}}

\usepackage{cite}
\usepackage{amsmath,amssymb,amsfonts}
\usepackage{algorithmic}
\usepackage{graphicx}
\usepackage{textcomp}
\usepackage{xcolor}
\usepackage{multirow}
\usepackage{subfigure}
\usepackage{array}
\begin{document}

\title{BASM: A Bottom-up Adaptive Spatiotemporal Model for Online Food Ordering Service}

\author{

\IEEEauthorblockN{Boya Du, Shaochuan Lin, Jiong Gao, Xiyu Ji,
        Mengya Wang,
        Taotao Zhou,
        Hengxu He\textsuperscript{*},
        Jia Jia\textsuperscript{*},
        Ning Hu}
\IEEEauthorblockA{
\textit{Alibaba Group}\\
Hangzhou\&Shanghai, China \\
\{
Boya.dby,
lin.lsc,
jionggao.gg,
        jixiyu.jxy,
        mia.wmy,
        taotao.zhou,
        hengxu.hhx,
        jj229618,
     huning.hu
\}@alibaba-inc.com}
\thanks{ * Corresponding author}
}


\maketitle

\begin{abstract}

Online Food Ordering Service (OFOS) is a popular location-based service that helps people to order what you want. Compared with traditional e-commerce recommendation systems, users' interests may be diverse under different spatiotemporal contexts, leading to various spatiotemporal data distribution, which limits the fitting capacity of the model. However, numerous current works simply mix all samples to train a set of model parameters, which makes it difficult to capture the diversity in different spatiotemporal contexts. Therefore, we address this challenge by proposing a Bottom-up Adaptive Spatiotemporal Model(BASM) to adaptively fit the spatiotemporal data distribution, which further improve the fitting capability of the model. Specifically, a spatiotemporal-aware embedding layer performs weight adaptation on field granularity in feature embedding, to achieve the purpose of dynamically perceiving spatiotemporal contexts. Meanwhile, we propose a spatiotemporal semantic transformation layer to explicitly convert the concatenated input of the raw semantic to spatiotemporal semantic, which can further enhance the semantic representation under different spatiotemporal contexts. Furthermore, we introduce a novel spatiotemporal adaptive bias tower to capture diverse spatiotemporal bias, reducing the difficulty  to model spatiotemporal distinction. To further verify the effectiveness of BASM, we also novelly propose two new metrics, Time-period-wise AUC (TAUC) and City-wise AUC (CAUC). Extensive offline evaluations on public and industrial datasets are conducted to demonstrate the effectiveness of our proposed modle. The online A/B experiment also further illustrates the practicability of the model online service. This proposed method has now been implemented on the Ele.me, a major online food ordering platform in China, serving more than 100 million online users.
\end{abstract}


\section{Introduction}

Online Food Ordering Service (OFOS), a convenient location-based service for ordering, delivering, or picking up food via a smartphone or website, has become popular in recent years. Nowadays, online food ordering platform, such as Grubhub, DoorDash, Meituan and Ele.me, has served millions of users every day. For instance, in the U.S., users
place more than 400 million online orders on the DoorDash food ordering platform in Q2 of 2022\footnote{https://ir.doordash.com/financials/quarterly-results/default.aspx}.





\begin{figure}[tbp]
{\includegraphics[width=1.0\columnwidth]{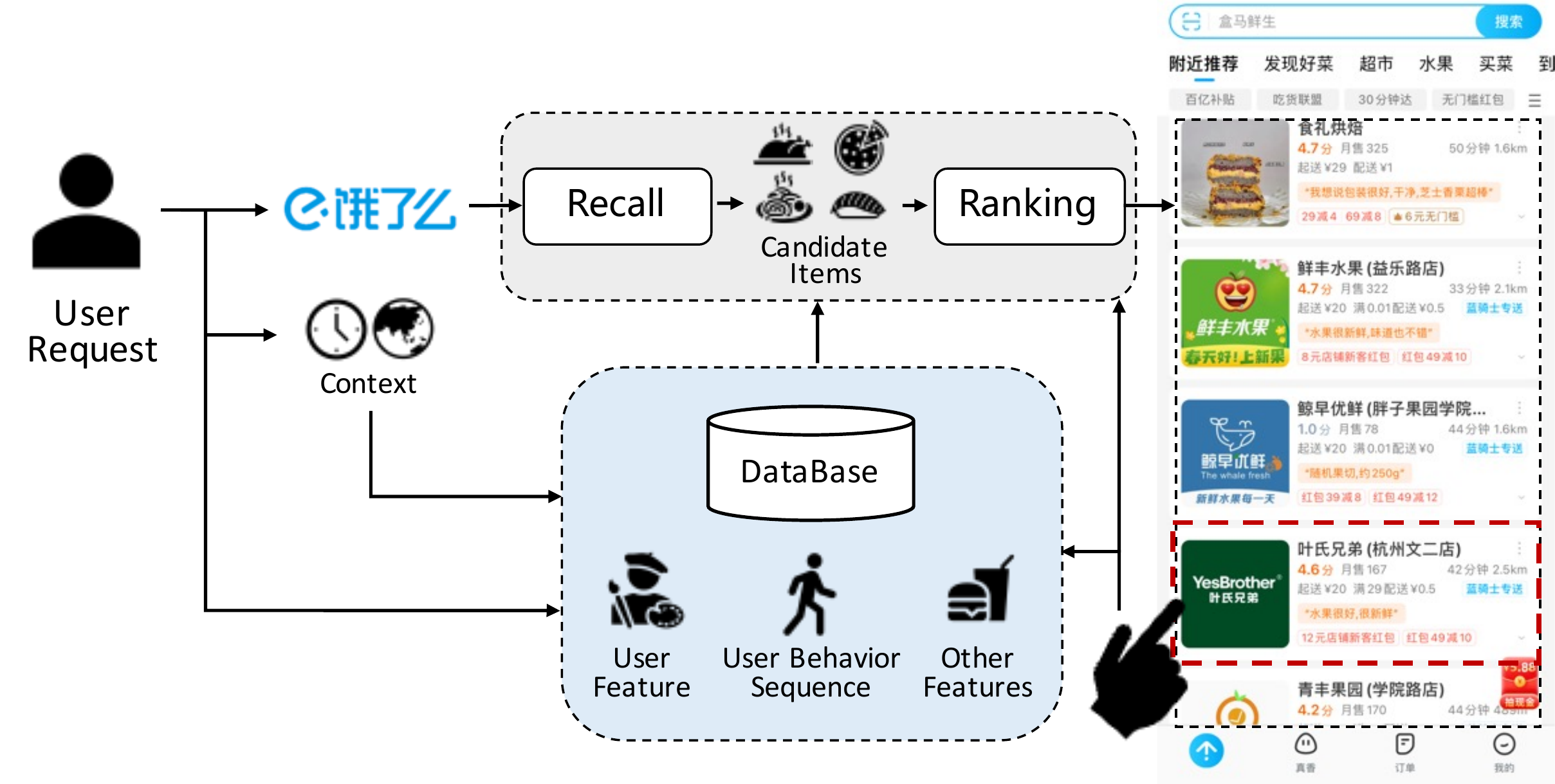}}
\caption{The overview of one typical online food ordering service implementation process.}
\label{fig:workflow}
\end{figure}

Compared with traditional e-commerce recommendation systems, a qualified OFOS needs to consider the user's request time and space to provide food ordering services. An example of one typical online food ordering service implementation process can be seen in Fig.~\ref{fig:workflow}. How to better predict user preferences under spatiotemporal constraints and continuously improving the performance of food recommender systems has become a heated topic in industry and academia. From the perspective of model parameters, the existing work on spatiotemporal data modeling can be divided into static parameter based methods \cite{autoint, TRISAN, ST-PIL, STAN, StEN, SIM, ETA, SDIM, SAM}  and dynamic parameter based methods \cite{ALL,STAR,AESM,ADI,ZEUS,DADNN,M2M,Lhuc,CAN,PTUPCDP,APG}. The static parameter based methods adopt a set of model parameters to fit entire data, and the dynamic parameter based methods exploit a set of adaptive model parameters or maintains multiple sets of model parameters at the same time to realize the dynamic modeling of data.

\begin{figure*}[tbp]
\centering
\subfigure[The distribution of exposures and CTRs over different hours.]{\includegraphics[width=1.0\columnwidth]{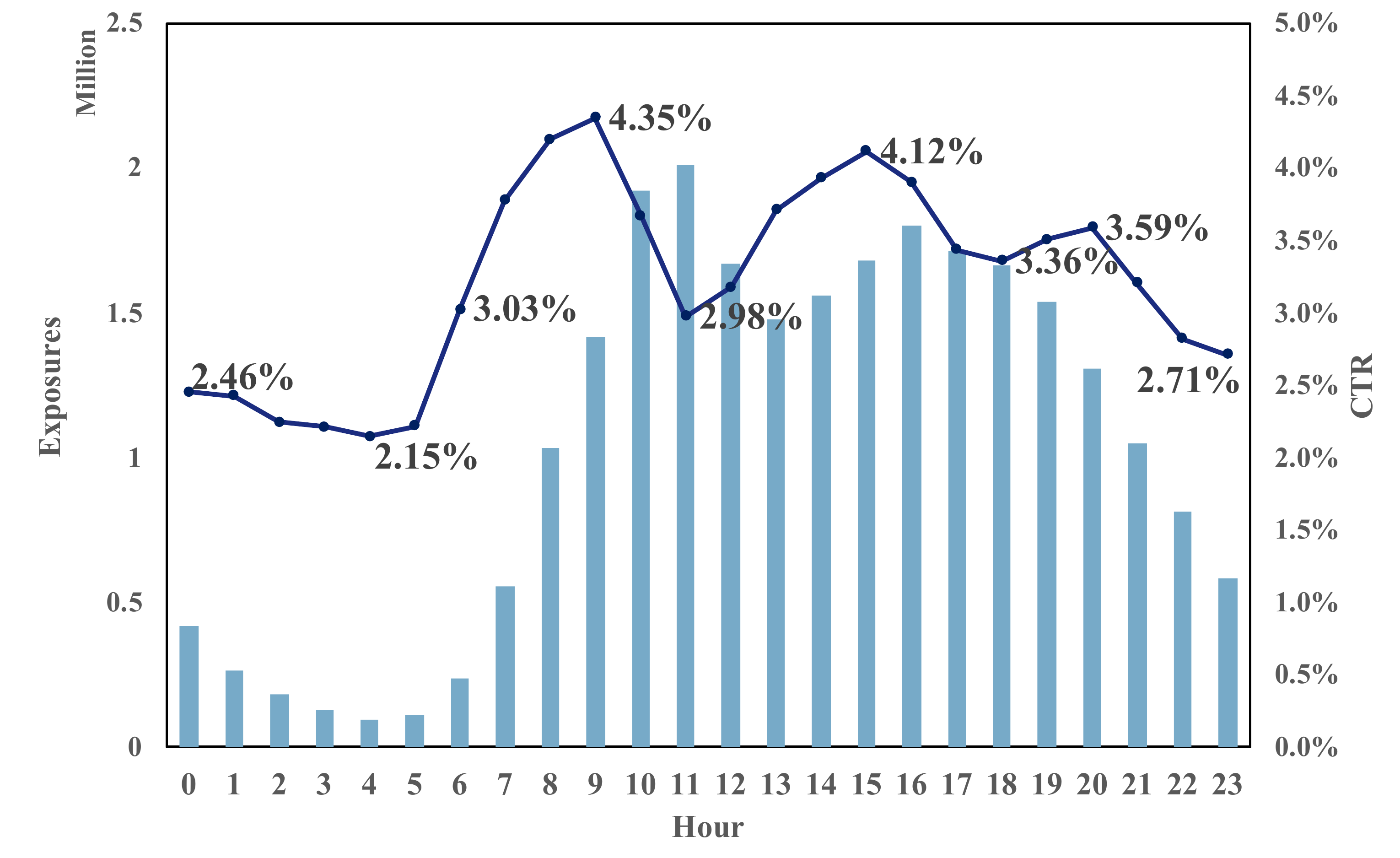}}
\subfigure[The distribution of exposures and CTRs over different cities.]{\includegraphics[width=1.0\columnwidth]{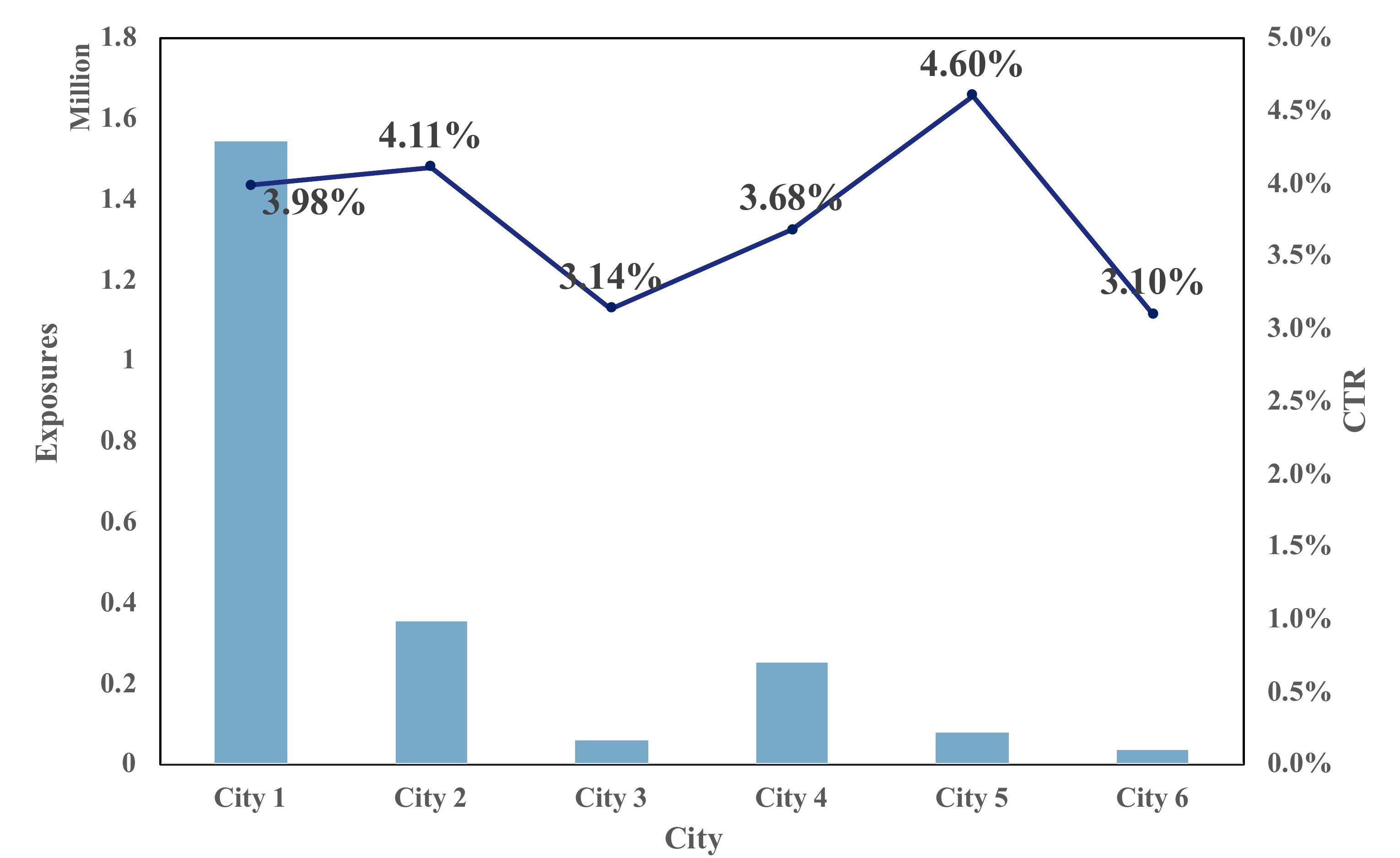}}
\caption{The distribution of exposures and CTRs across various spatiotemporal scenarios (i.e., hours and cities) for one week in Aug 2022.}
\label{fig:introduction}
\end{figure*}

For static parameter based methods, existing works mainly focus on extraction of spatiotemporal interest and enhancement of spatiotemporal information. Among them, the spatiotemporal periodic user behavior modeling methods \cite{ST-PIL, StEN} divides and combines user behavior sequences through spatiotemporal information, and realizes the extraction of users' periodic spatiotemporal interests. Long-sequence user behavior modeling methods\cite{SIM, ETA, SDIM, SAM} extracts users' long-term interest preferences by extending the length of user behavior sequences. TRISAN\cite{TRISAN} enhances the representation of spatiotemporal information by constructing a triangular relationship between user geographic location, item geographic location, and user behavior click time. AutoInt\cite{autoint} enriches the representation of input information by automatically  features interactions from multiple subspaces.

For dynamic parameter based methods, the adaptive model parameter methods \cite{M2M,Lhuc,CAN,PTUPCDP,APG} achieve adaptive learning of the model by maintaining a set of dynamic parameters. M2M\cite{M2M} constructs a meta-unit to generate tower weights in different scenarios to realize dynamic modeling of multi-scenario data. APG\cite{APG} realizes self-wise adaptive modeling by building parameter generation network and parameter adaptor. The multi-parameter learning methods \cite{ALL,STAR,AESM,ADI,ZEUS,DADNN} achieve fitting of different data distributions by maintaining multiple sets of model parameters at the same time. STAR\cite{STAR} realizes adaptive modeling of multi-scene data distribution by constructing shared parameters and domain-specific parameters. ADI\cite{ADI} further optimizes the fusion mechanism between shared parameters and domain-specific parameters to improve the feature-level domain adaptation.

However, in online food ordering services, users have different preferences in different spatiotemporal contexts, resulting in large differences in spatiotemporal data distribution. Taking the Ele.me recommendation paltform as an example, as shown in Fig.~\ref{fig:introduction}, the data distribution (exposure and CTR) would be varied at different time ($e.g.$, 6 a.m vs 6 p.m) and at different location ($e.g.$, city 1 vs city 2). Static parameter based methods simply mix all samples to train a set of model parameters, which makes it difficult to capture diversities in different spatiotemporal contexts. Meanwhile, dynamic parameter based methods either ignore the effects of spatiotemporal factors or rely heavily on predefined parameter spaces, such as scenarios and crowds. In OFOS, spatiotemporal factors, as one of the dominant factors in user decision-making, play an important role in spatiotemporal data modeling. Moreover, the spatiotemporal scenario is continuous and non-enumerable, and it is difficult to predefine different spatiotemporal parameter spaces.

To echo the above challenge, which we name \textit{spatiotemporal data distribution}, we propose \textbf{B}ottom-up \textbf{A}daptive \textbf{S}patiotemporal \textbf{M}odel, \textbf{BASM} for short, adaptive modeling of spatiotemporal data from the bottom embedding layer, middle semantic layer and top classification tower. Specifically, at the bottom of the information input, we propose the  Spatiotemporal-Aware Embedding Layer, which performs weight adaptation from the field granularity to achieve the purpose of dynamically perceiving spatiotemporal contexts. In this case, the spatiotemporal weights of features are dynamically scaled as the spatiotemporal context changes. 
Concatenating the raw semantic inputs obtained from the above aware features, we further propose a Spatiotemporal Semantic Transformation Layer. The raw semantic is explicitly converted to spatiotemporal semantic to further enhance the semantic representation under different spatiotemporal contexts. Meanwhile, we novelly propose a Spatiotemporal Adaptive Bias Tower, which modulates the parameters of raw fully-connected layer and batch normalization layer through spatiotemporal information to realize the adaptive modeling of the spatiotemporal bias, thereby reducing the difficulty of spatiotemporal distinction modeling. To further verify the effectiveness of  BASM, we also propose two new metrics, time-period-wise AUC (TAUC) and city-wise AUC (CAUC), which can be seen in Section.~\ref{experiment}. Comprehensive experiments on public and industrial datasets demonstrate the superiority of our proposed method compared to state-of-the-art methods, and online A/B test illustrates effectiveness and efficiency of BASM.






In summary, the main contributions of this work can be listed as follow:

\begin{itemize}
    \item We novelly propose a Bottom-up Adaptive Spatiotemporal Model(BASM) to characterize various spatiotemporal distribution, which further improve the fitting capability of the model. 
    \item We realize adaptive modeling of spatiotemporal data from the bottom embedding layer, middle semantic layer and top classification tower by proposing the Spatiotemporal-Aware Embedding Layer, the Spatiotemporal Semantic Transformation Layer and the Spatiotemporal Adaptive Bias Tower. 
    \item To further verify the effectiveness of BASM, we also propose two new metrics, time-period-wise AUC (TAUC) and city-wise AUC (CAUC), which are demonstrated in Section.~\ref{experiment}.
    \item Extensive experiments on industrial dataset with more than 80 million users and a large-scale public dataset demonstrate the effectiveness of BASM, further validated by online A/B tests on industrial spatiotemporal scenarios. BASM has now been successfully deployed on Ele.me, one of the major online food ordering platform in China, achieving a CTR improvement of 6.51\% up to now.
\end{itemize}


\begin{figure*}[tp]
  \centering
  {
      \includegraphics[width=1.8\columnwidth]{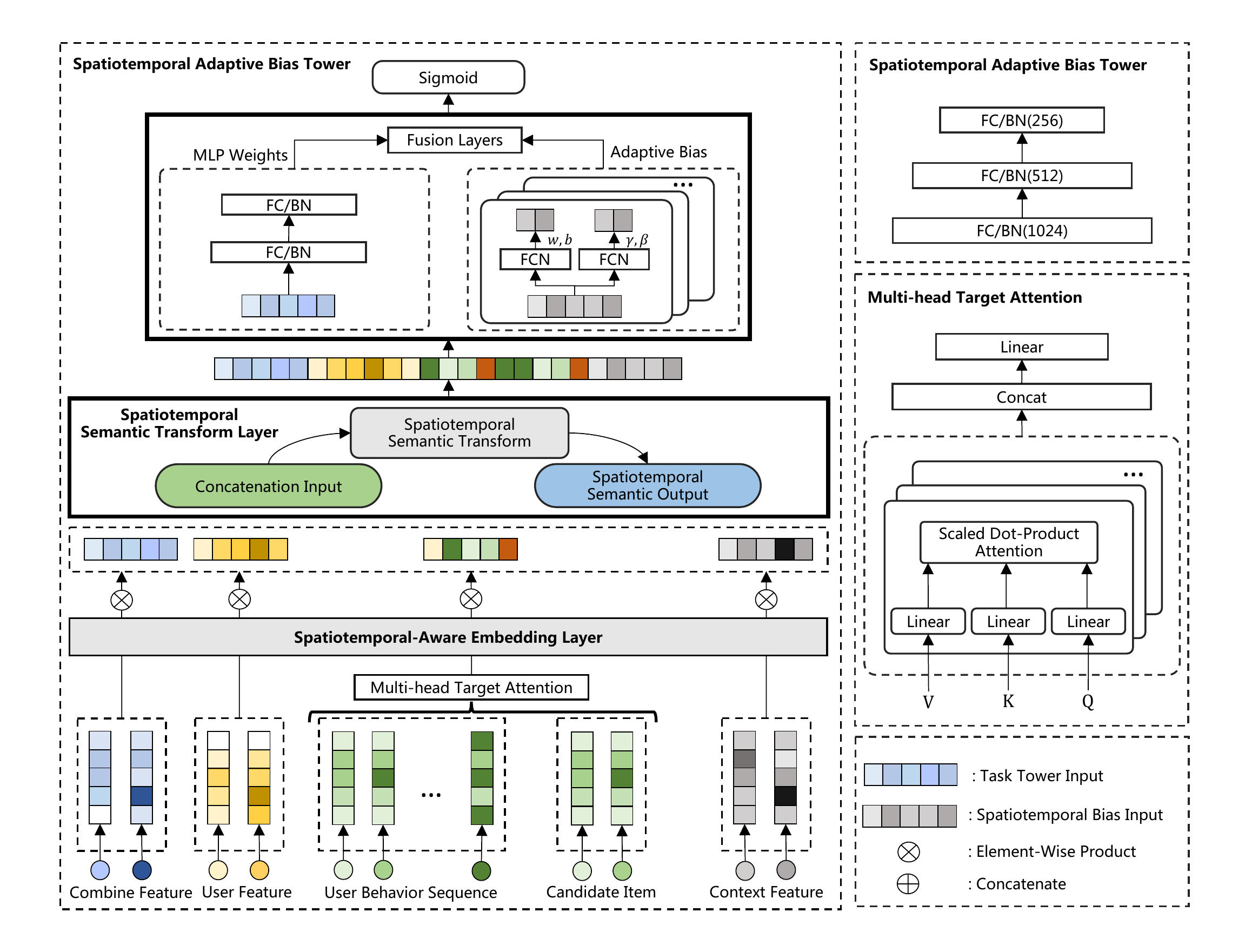}
  }
  \caption{Bottom-up Adaptive Spatiotemporal Model (BASM) architecture overview. It consists of three modules: the Spatiotemporal-Aware Embedding Layer (described in Section~\ref{sec:StAEL}), the Spatiotemporal Semantic Transformation Layer (described in Section~\ref{sec:StSTL}) and the Spatiotemporal Adaptive Bias Tower (described in Section~\ref{sec:StABT}).}
  \label{fig:framework}
\end{figure*}

\section{Proposed Method}
In this section, we introduce the problem of spatiotemporal data distribution and detail the Bottom-up Adaptive Spatiotemporal Model (BASM), with its overall architecture illustrated in Fig.~\ref{fig:framework}. BASM proposes the Spatiotemporal-Aware Embedding Layer, the Spatiotemporal Semantic Transformation Layer and the Spatiotemporal Adaptive Bias Tower, which comprehensively solves the above problems from the bottom embedding layer, the middle semantic layer and the top classification tower, respectively.

\begin{table}[htbp]
    \centering
    \caption{Detailed Features contained in each Field}\label{tab:FeaField}
    \begin{tabular}{c|c}
    \hline
        \makebox[2.5cm][c]{\textbf{Field}} & \makebox[5cm][c]{\textbf{Detailed Features}} \\
    \hline
        \multirow{2}*{User Feature} & User ID, Basic Profiles, Statistics of User's \\
        ~&Ordering / Clicking / Exposure / etc. \\
    \hline
        \multirow{2}*{User Behavior Sequence} & Item ID, Brand, General Category, Time-\\
        ~&Period, Hour, City ID, and etc. \\
    \hline
        \multirow{3}*{Candidate Item} & Item ID, General Category, Brand, Position, \\
        ~&Statistics of Shop's Ordering / Clicking / \\
        ~&Exposure / etc. \\
    \hline
        Spatiotemporal & \multirow{2}*{Time-period / Hour / Geohash/ City ID / etc.}  \\
        Context Feature&~\\
    \hline
        \multirow{2}*{Combine Feature} & Some Hand-selected Cross-features \\
        ~&between Users and Items.\\
    \hline
    \end{tabular}
    \label{tab:field}
\end{table}

\begin{table}[htbp]
    \centering
    \caption{The notation used in this paper}\label{tab:Denotion}
    \begin{tabular}{c|c}
    \hline
        \textbf{Symbol} & \makebox[5cm][c]{\textbf{Description}} \\
    \hline
        $N$ & The number of all unique features. \\
    \hline
        $n$ & The number of feature fields. \\
    \hline
        $c$ & The spatiotemporal context feature. \\
    \hline
        \multirow{2}*{$r$} & All other features except spatiotemporal\\
        ~&context features. \\
    \hline
        ${e_i}$ & Embedding vector of the $i^{th}$ feature. \\
    \hline
        \multirow{2}*{${h_j}$} & Hidden representation of the $j^{th}$ \\ ~&feature field. \\
    \hline
        \multirow{2}*{${h_c}$} & Hidden representation of the spatiotemporal \\
        ~&context feature field. \\
    \hline
        \multirow{2}*{${\alpha_j}$} & Spatiotemporal weight  of the $j^{th}$ feature \\
        ~&field learning by StAEL. \\
    \hline
        \multirow{2}*{${W_{stl}, b_{stl}}$} & Dynamic parameter weights and bias \\
        ~&learned by StSTL. \\
    \hline
        \multirow{2}*{$W_{bias}^{(m-1)},b_{bias}^{(m-1)}$} & Dynamic bias of FCNs learning by \\
        ~&Spatiotemporal Adaptive Bias Tower. \\
    \hline
        \multirow{2}*{${\gamma_{bias}^{{(m-1)}}, \beta_{bias}^{{(m-1)}}}$} & Dynamic bias of BNs learning by \\
        ~&Spatiotemporal Adaptive Bias Tower. \\
    \hline
        $y_i, {\hat{y}_i}$ & The click label and prediction of instance $i$. \\
    \hline
        $\sigma(\cdot)$ & A non-linear activation function.  \\
    \hline
        $[;]$ & Concatenation operator.\\
    \hline
    \end{tabular}
\end{table}

\subsection{Problem Formulation}
In this paper, to better maintain and manage the features in our application, we classify the entire input features and declare each class as a field. As shown in TABLE ~\ref{tab:field}, let $\mathcal{X}$ represents the input data, which consists of the spatiotemporal context feature field $c$, the user feature field, the user behavior sequence field, the candidate feature field and the combine feature field. To better observe the effects of spatiotemporal features, in the following, we refer to other features except spatiotemporal context features as $o$, which is $x_i=(o_i, c_i)$, where $x_i\in\mathcal{X}$ and $i$ represent the instance. 
$\mathcal{Y}$ denotes the click label. Formally, the \textit{spatiotemporal data distribution} problem can be formulated by,
\begin{align}
    y_i &= \mathcal{F}_{\theta}(x_i) \\
    &= \mathcal{F}_{\theta}(o_i, c_i)
\label{eq:ctr_goal}
\end{align}
where $y_i \in \mathcal{Y}$, $\mathcal{F}$ is the deep network and $\theta$ is the parameters of the deep network. Different from the e-commerce scenario, spatiotemporal context features $c$, as one of the main features of online order recommendation systems, play an important role in spatiotemporal data modeling. Moreover, the spatiotemporal scenario is continuous and non-enumerable, and it is difficult to predefine different spatiotemporal parameter spaces.
Therefore, we propose to solve the spatiotemporal data distribution problem by  dynamically adapt the parameter $\theta$ under various spatiotemporal factors $c$. Below we will introduce our model module by module.


\subsection{Spatiotemporal-Aware Embedding Layer (StAEL)}
\label{sec:StAEL}



\begin{figure}[tbp]
\centerline{\includegraphics[width=1.0\columnwidth]{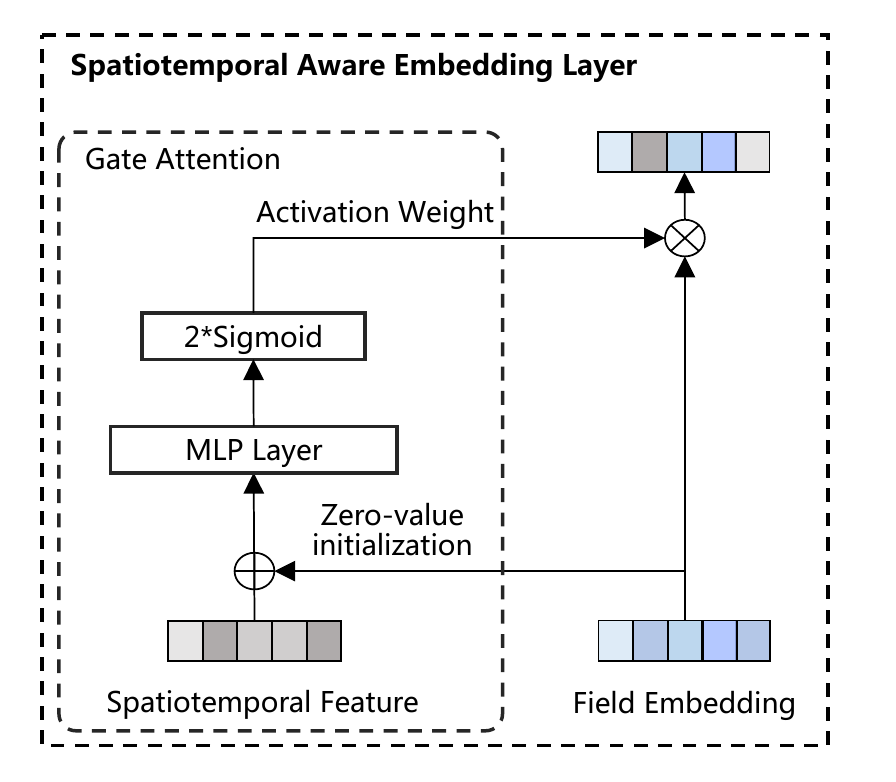}}
\caption{The architecture of Spatiotemporal-Aware Embedding Layer. It applies a gate attention mechanism to perform weight adaptation from the field granularity.}
\label{fig:SAEL}
\end{figure}

As with other deep models, we first encode each discrete class feature into a high-dimensional one-hot vector. For the $i^{th}$ feature, its one-hot encoding is denoted as:
\begin{equation}
    {v_i} = onehot(i)
    \label{eq:sample}
\end{equation}
where ${v_i} \in \mathbb{R}^N$ is a vector with 1 at the $i^{th}$ entry and 0 elsewhere, and N is the number of all unique features. We then embed the sparse and high-dimensional one-hot encoding vectors to dense and low-dimensional vectors, which are more suitable for neural network inputs. In particular, we define a learnable embedding matrix ${E} \in \mathbb{R}^{D\times N}$ and project the $i^{th}$ feature to its
corresponding embedding vector:
\begin{equation}
    {e_i = E v_i}
    \label{eq:sample}
\end{equation}
where ${e_i} \in \mathbb{R}^{D}$, $D$ is the dimension, much smaller than $N$.

Most of the existing methods have the same feature importance in different spatiotemporal scenarios, ignore the distribution differences of spatiotemporal data, and fail to achieve feature weight adaptation, which limits the model's perception of spatiotemporal contexts. For example, during mealtimes, users are more concerned about the price of food, so the price feature will be more important. In contrast, during the time-period of afternoon tea, users are more inclined to browse around and compare multiple shops, so the category feature of the shop will be more concerned. Uniform feature weights cannot dynamically perceive such spatiotemporal changes. However, it is impossible to maintain a set of embedding for each spatiotemporal scenario. Because this method not only needs to maintain a large number of parameters, but also is difficult to fit the few-shot distribution. Typically, due to the large number $N$ of unique features in industrial recommendation systems, more than hundreds, it is prohibitively expensive to learn different importance weights for each feature. This not only brings a huge amount of computation, but also reduces the fitting effect of the model due to too much noise.



Therefore, we dynamically characterize each field-wise feature embeddings according to the spatiotemporal context feature $c$, and each field representation is defined as:

\begin{equation}
    {h_j} = \alpha_{j} {x_j}
    \label{eq:hi}
\end{equation} 
where ${x_j}=[{e_{1j};\dots; e_{kj}}]$ denotes the $j^{th}$ field embedding, which is the concatenation of all feature embeddings in it. $k$ is the total number of features in the $j^{th}$ field. $\alpha_j$ is a spatiotemporal weight assigned to ${x_j}$, indicating its importance in current spatiotemporal context.  $\{j\in \mathbb{Z}: 0 \leq j < n\}$, where $n$ denotes the number of feature fields. ${h_j}$ denotes the adaptive embedding of the $j^{th}$ field of features.


The aim of our Spatiotemporal-Aware Embedding Layer is to calculate the appropriate spatiotemporal weight $\alpha_j$. A naive way is $\alpha_{j} = 1/n$, that is, each feature field shares the same weight. This is obviously not a wise choice,  because the importance of features cannot be adapted to various spatiotemporal contexts. Illustrated in Fig.~\ref{fig:SAEL}, we apply a gate attention mechanism to extract the spatiotemporal weight over different spatiotemporal contexts, which can be seen as follows:

\begin{equation}
    \alpha_{j} = 2 * \sigma{ \left( {{W_p}} [{x_j}; {x_{c}}] + b_{p} \right)}
    \label{eq:sample}
\end{equation}
where, $\sigma$ denotes the activation function, which is the Sigmoid function in our application. ${x_j}$ and ${x_{c}}$ represents the embeddings of the other feature field and spatiotemporal context features, respectively. $[;]$ is the concatenation operator, and $W_p$ and $b_{p}$ are attention parameters. The spatiotemporal weight is activated by the Sigmoid function and then multiplied by 2 to scale the weight between 0 and 2. The purpose of this is mainly to ensure that different features are strengthened ($i.e.$, $>1$) or weakened ($i.e.$, $<1$) under different spatiotemporal contexts. Through adopting such weight adaptation method, we can achieve the purpose of dynamically perceiving spatiotemporal contexts at the bottom layer. 



\subsection{Spatiotemporal Semantic Transformation Layer (StSTL)}
\label{sec:StSTL}


\begin{figure}[tbp]
\centerline{\includegraphics[width=1.0\columnwidth]{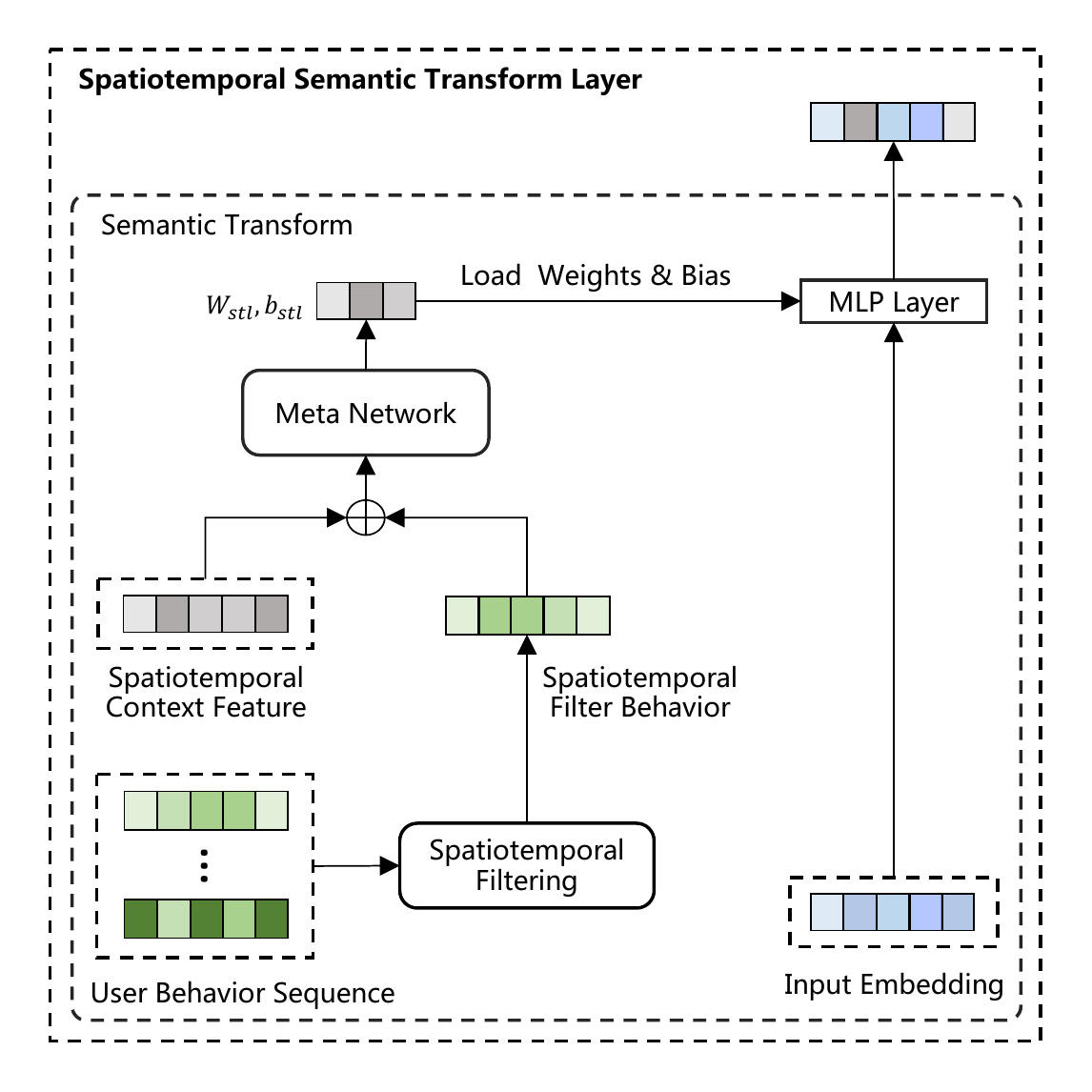}}
\caption{The architecture of Spatiotemporal Semantic Transformation Layer. It utilizes a meta network to explicitly transform the raw semantic into the spatiotemporal semantic.}
\label{fig:SSTL}
\end{figure}


After obtaining the dynamic spatiotemporal embedding by the above aware layer, we can obtain raw semantic by a simple concatenation  ${\hat{h} = [h_{0} ;\dots;h_{n-1} ]}$. However, this raw semantic is difficult to characterize the dinstinction under  different spatiotemporal contexts, which limits the fitting capability of the model to the spatiotemporal data. Hence, we propose to explicitly transform the raw semantic into the spatiotemporal semantic by bringing dynamic network parameters,  illustrated in Fig.~\ref{fig:SSTL}.


Inspired by \cite{CAN, M2M}, a meta network is utilized to achieve semantic transformation purpose. Since the raw semantic is closely related to the user, in addition to the spatiotemporal context, we further bring the historical behavior of the user side under this spatiotemporal context to improve the representation ability of spatiotemporal dynamic parameters. Specially, we exploit the  time-period and geohash to filter the user's historical behavior,  so as to obtain the personalized spatiotemporal filtering behavior $ui$. Then, we concatenate the above-mentioned spatiotemporal context embedding ${h_{c}}$ and spatiotemporal filtering behavior embedding ${h_{ui}}$ to obtain dynamic parameter weights ${W_{stl}}$ and dynamic parameters bias ${b_{stl}}$ through a meta network,



\begin{align}
    {W}_{stl} & = \operatorname{Reshape}\left( {{W_{w}}} [{h_{c}}; {h_{ui}}]+{b_{w}} \right)    \\
    {b}_{stl} & = \operatorname{Reshape}\left( {{W_{b}}} [{h_{c}}; {h_{ui}}]+{b_{b}} \right)  
\end{align}
where $\operatorname{Reshape}$ is used to convert the output embedding of the meta network into a dynamic weight matrix. ${W_{w}}$ and ${b_{w}}$ are the weight and bias of meta network, so as ${W_{b}}$ and ${b_{b}}$. ${W_{stl}}$ and ${b_{b}}$ are the dynamic weight and dynamic bias from meta network output.


After generating the dynamic weight matrix and bias vector, the raw input  semantic ${\hat{h}}$  is transformed into spatiotemporal semantic output,

\begin{align}
   {h_{stl}^{*}} = {{W}_{stl}} {\hat{h}} + {b}_{stl}
\end{align}
where ${h_{stl}^{*}}$ is the spatiotemporal semantic output. Different from traditional linear mapping methods that lack spatiotemporal adaptation, we implement different mapping functions under different spatiotemporal contexts through a meta-learning network. In this way we achieve an explicit transformation of raw semantic to spatiotemporal semantic, enhancing the semantic representation of different spatiotemporal contexts.


\subsection{Spatiotemporal Adaptive Bias Tower (StABT)}
\label{sec:StABT}

\begin{figure}[tp]
\centerline{\includegraphics[width=0.9\columnwidth]{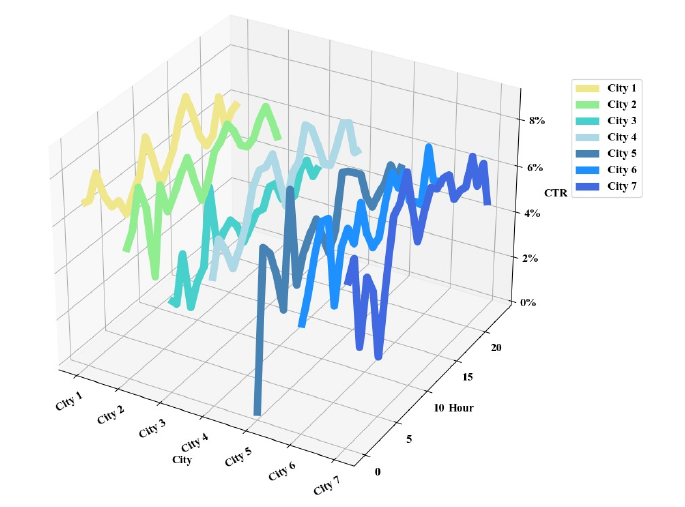}}
\caption{The spatiotemporal bias over different cities and hours.}
\label{fig:time_city_3d}
\end{figure}

\begin{figure}[tbp]
\centerline{\includegraphics[width=1.0\columnwidth]{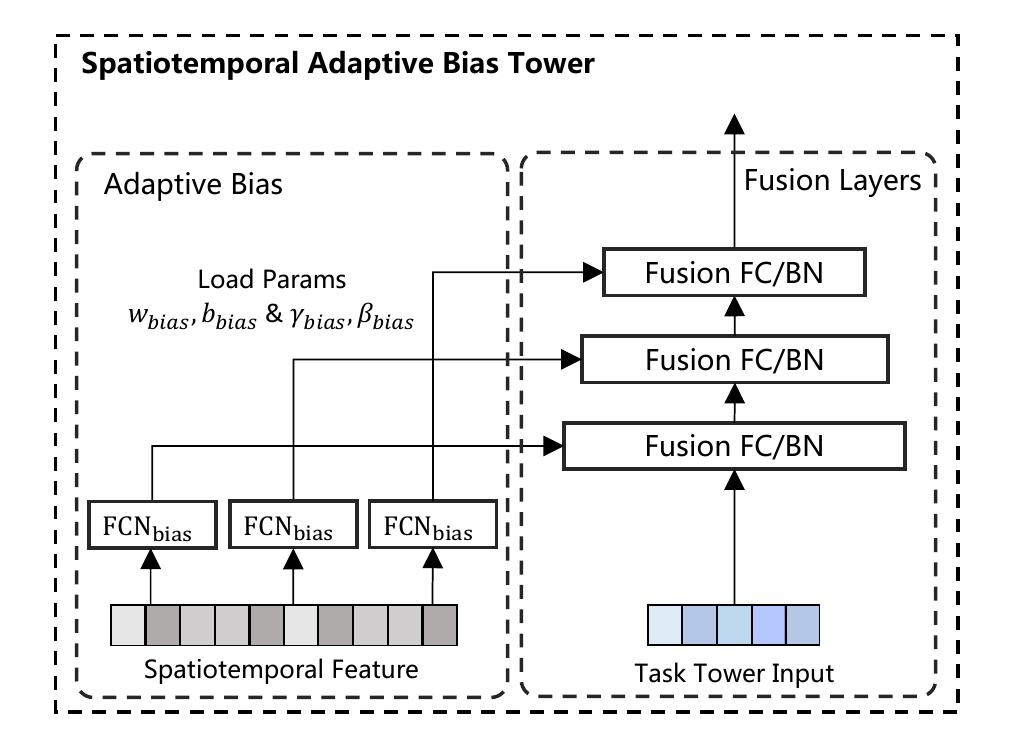}}
\caption{The architecture of Spatiotemporal Adaptive Bias Tower. It consists of Fusion FCs and Fusion BNs.}
\label{fig:StMT}
\end{figure}

In OFOS, there are natural differences in CTR under different time and location, and we call this phenomenon spatiotemporal bias. The spatiotemporal bias can be seen in Fig.~\ref{fig:time_city_3d}, we can find that in different time and location, the user's click tendency is various, which cannot be ignored in the spatiotemporal data modeling. As shown in Fig.~\ref{fig:StMT}, in order to capture diverse spatiotemporal bias given different spatiotemporal scenarios, we novelly present a Spatiotemporal Adaptive Bias Tower, which consists of fusion Fully Connected layers (FCs) and fusion Batch Normalization layers (BNs).



\textbf{Fusion FCs:} Static parameters based FCs are easily dominated by the \textit{strong} distribution and will cause the \textit{weak} distribution to be submerged. Therefore, we exploit spatiotemporal context features to construct spatiotemporally specific parameters to modulate the weights of FCs to better characterize spatiotemporal bias given different spatiotemporal scenarios. The spatiotemporal dynamic parameters of FCs modulation can be obtained as follows:


\begin{align}
{W_{bias}^{{(m-1)}}}  = \sigma &\left({W_{dfc,1}^{(m-1)}} {h_{c}}+{b}_{dfc,1}^{(m-1)}\right)    \label{eq:wbias} \\
{b_{bias}^{{(m-1)}}}  = \sigma &\left({W_{dfc,2}^{(m-1)}} {h_{c}}+{b}_{dfc,2}^{(m-1)}\right)   \label{eq:bbias}
\end{align}
where ${W_{bias}^{{(m-1)}}}$ and ${b_{bias}^{{(m-1)}}}$ are the spatiotemporal modulation parameters for FC.  ${W_{dfc,1}^{(m-1)}}$, ${b}_{dfc,1}^{(m-1)}$, ${W_{dfc,2}^{(m-1)}}$ and ${b}_{dfc,2}^{(m-1)}$ are the network weights that output this modulation parameters. $\{m \in \mathbb{Z}: 1 \leq j \leq L\}$, and $L$ is the layers of the tower. Our dynamic bias modulation for FCs is obtained by:

\begin{equation}
h_{fusion}^{(0)} = {h_{stl}^{*}}
\end{equation}
\begin{equation}
\begin{aligned}
\begin{split}
h_{fusion}^{(m)} =\sigma \left( \left({W_{bias}^{(m-1)}} \odot {W_t^{(m-1)}} \right) h_{fusion}^{(m-1)} + \right. \\ 
\left. \left({b}_{bias}^{(m-1)} + {b}_t^{(m-1)}\right) \right) , \forall m \in 1,2, .., L
\end{split}
\end{aligned}
\end{equation}
where $\odot$ is Hadamard product, ${W_t^{(m-1)}}$ and ${b}_t^{(m-1)}$ are the MLP weights to be modulated. By this means, we implement Fusion FCs, which enables dynamic modeling under spatiotemporal bias by modulating the weights of FCs.


\textbf{Fusion BNs:} Batch  normalization$(\mathrm{BN})$ has been proven to be effective and widely applied in deep learning. Let $\mu$ denotes the mean value of input $\mathcal{X}$, while $\sigma^2$ denotes the variance, the raw BN can be shown as follows:

\begin{align}
\hat{\mathcal{X}}=\gamma \frac{\mathcal{X}-\mu}{\sqrt{\sigma^2+\epsilon}}+\beta
\end{align}
where $\gamma$ and $\beta$ are learnable parameters, $\epsilon$ is a small number to avoid the denominator being 0. Raw $\mathrm{BN}$ requires the input $\mathcal{X}$ to satisfy the assumption of independent and identical distribution(i.i.d), which works well in single scenario. However, in our spatiotemporal scenario, the data distribution corresponding to each spatiotemporal context is diverse. The raw BN confuses the distributional differences across different spatiotemporal contexts with statistics from all samples. Therefore, we need to utilize the spatiotemporal features to modulate the raw BN.




\begin{align}
{\gamma_{bias}^{{(m-1)}}}  &= \sigma \left({W}_{dbn,1}^{(m-1)} h_c + {b}_{dbn,1}^{(m-1)}\right)    \label{eq:gamma} \\
    {\beta_{bias}^{{(m-1)}}} &= \sigma \left({W}_{dbn,2}^{(m-1)} h_c+{b}_{dbn,2}^{(m-1)}\right) \label{eq:beta} \\
{\hat{\mathcal{X}}_{fusion}^{(m)}} =  \gamma_{bias}^{(m-1)} & \gamma^{(m-1)}  \frac{{\mathcal{X}^{(m-1)}}-\mu}{\sqrt{\sigma^2+\epsilon}} +  \beta^{(m-1)} + \beta_{bias}^{(m-1)} 
\end{align}
where ${W}_{dbn,1}^{(m-1)}$, ${b}_{dbn,1}^{(m-1)}$, ${W}_{dbn,2}^{(m-1)}$ and ${b}_{dbn,2}^{(m-1)}$ are also the four network weights. Eq.~\ref{eq:wbias}, Eq.~\ref{eq:bbias}, Eq.~\ref{eq:gamma} and Eq.~\ref{eq:beta} are collectively named $\mathrm{FCN}_{bias}$ in Fig.~\ref{fig:StMT}. ${\gamma_{bias}^{{(m-1)}}}$ and ${\beta_{bias}^{{(m-1)}}}$ represent modulated parameters. ${\gamma^{{(m-1)}}}$ and ${\beta^{{(m-1)}}}$ denote the origin learnable $\mathrm{BN}$ parameters. By the above approach, we achieve fusion BNs by modulating the parameters of BNs, and realize the dynamic normalization in different spatiotemporal data distributions. Experiments show that this method is more beneficial for spatiotemporal data modeling.


After Spatiotemporal Adaptive Bias Tower, our final click prediction can be presented as:
\begin{align}
    \hat{y}_i = \sigma \left( W_o \hat{x}_i + b_o \right)
\end{align}
where, $\hat{x}_i$ represents the final output of Spatiotemporal Adaptive Bias Tower of instance $i$. Then we update the entire network by minimizing the binary cross-entropy loss with the prediction $\hat{y}_i$ and the label ${y}_i$:
\begin{equation}
     \mathop{min}\limits_{\theta}\sum_{i}-y_{i}log\hat{y}_i-(1-y_i)log(1-\hat{y}_i)
\end{equation}

\begin{table*}[htbp]
\centering
  \caption{the basic Statistics of datasets. \# denotes the number, and ML indicates the average length of user behavior sequences.}\label{tab:datasetS}
  \begin{tabular}{ccccccc}
  \hline
    Datasets&Total Size&{\#Feature}&{\#Users}&{\#Items}&{\#Clicks}&{ML of User Behaviors} \\
  \hline
    \emph{Ele.me}&2380427866&417&81086293&547354&86735276&42.86 \\
    \emph{Spatiotemporal Public Data}&177114244&38&14427689&7446116&3140831&41.19 \\
  \hline
  \end{tabular}
 \end{table*}

\begin{table*}[htbp]
\centering
  \caption{Offline Performance Comparison with static and dynamic parameter State-of-the-arts on public and industrial datasets. the best results are in bold. }\label{tab:PerComp}
  \begin{tabular}{c|cccccc|cccccc}
  \hline
    \multirow{2}*{Methods} &\multicolumn{6}{c}{Ele.me} \vline &\multicolumn{6}{c}{Spatiotemporal Public Data} \\ 
    \cline{2-13}
    ~&{AUC}&{TAUC}&{CAUC}&{NDCG3}&{NDCG10}&{Logloss}&{AUC}&{TAUC}&{CAUC}&{NDCG3}&{NDCG10} &{Logloss}\\
  \hline
    {Wide\&Deep} &0.7037& 0.7022 & 0.7016 & 0.8219 & 0.8410 & 0.1376 
    &0.5940 & 0.5920 & 0.6559 & 0.2231 & 0.3883 & 0.0923 \\
    {DIN} &0.7327& 0.7313 & 0.7302 & 0.8309 & 0.8482 & 0.1376
    & 0.6494 & 0.6382 & 0.6359 & 0.2806 & 0.4512 & 0.0843\\
    {AutoInt} & 0.7308 & 0.7295 & 0.7285 & 0.8272 & 0.8502 & 0.1346
    &0.6610 & 0.6583 & 0.6559 & 0.2781 & 0.4340 & 0.0842\\
    \hline
    {STAR} &0.7331&0.7321&0.7265&  0.8300 & 0.8523 & 0.1343 
    &0.6671 & 0.6637 & 0.6614 & 0.2993 & 0.4553& 0.0844\\
    {M2M} &0.7333&0.7319&0.7308&  0.8296 & 0.8514 & 0.1345 
    &0.6686& 0.6647 & 0.6634 & 0.2922 & 0.4440 & 0.0847\\
    {APG} & 0.7339 & 0.7326 & 0.7315 & 0.8301 & 0.8519 & 0.1344 
    &0.6634 & 0.6597 & 0.6579 & 0.2958 & 0.4508 & 0.0833\\
    \hline
    \textbf{BASM}&\textbf{0.7373}&\textbf{0.7360}&\textbf{0.7348} & \textbf{0.8323}& \textbf{0.8535}&\textbf{0.1339}
    &\textbf{0.6732} &\textbf{0.6686} &\textbf{0.6659} &\textbf{0.3240} & \textbf{0.4820} &\textbf{0.0827}\\
  \hline
  \end{tabular}
 \end{table*}

\section{Experiment}
\label{experiment}
In this section, we evaluate both the offline model and the
whole online system. Extensive comparisons and ablation studies are
conducted to answer the following questions:
\begin{itemize}
    \item \textbf{RQ1.} How does our proposed BASM
performs compared to state-of-the-art methods?
    \item \textbf{RQ2.} Where does the performance gain come from?
    \item \textbf{RQ3.} How efficient are BASM and comparison methods?
    \item \textbf{RQ4.} How does BASM perform on the online platform?
    \item \textbf{RQ5.} Can BASM cope with the before-mentioned challenges in the spatiotemporal recommendation scenario?
\end{itemize}
\subsection{Experimental Settings}

\subsubsection{Datasets}
To validate the performance of BASM, a takeaway industrial dataset \emph{Ele.me} and a public spatiotemporal recommendation dataset \emph{Spatiotemporal Public Data}\footnote{https://tianchi.aliyun.com/dataset/dataDetail?dataId=131047} are selected.
(1) \textbf{Ele.me:} \emph{Ele.me} is collected from the industrial recommendation platform, Ele.me. It has more than 80 million users and 2 billion samples. 45-day samples are used for training and the samples of the following day are used for testing. (2) \textbf{Spatiotemporal Public Data:} It contains a total of over 170 million data for 8 days. This dataset includes 7-day training data and the following 1-day test data. The statistics of the datasets are shown in Table \ref{tab:datasetS}.

\subsubsection{Comparison Methods}
We select three representative static parameter based methods and three dynamic parameter based methods for comparison to verify the effectiveness of our method on spatiotemporal data.
(1) \textbf{Wide\&Deep\cite{WDL}:} {Wide\&Deep} is an classic recommendation model of feature interactions. It jointly trains wide linear model for memorization and deep neural network for generalization. 
(2) \textbf{DIN\cite{zhou2018deep}:} DIN concentrates the diversity and local activation of user interests. A local activation unit extracts user interests from their extensive history behaviors.
(3) \textbf{AutoInt\cite{autoint}:} AutoInt utilizes multiple layers of the multi-head self-attention neural networks to simulate various orders of feature combinations, which can automatically learn the high-order feature interactions.
(4) \textbf{STAR\cite{sheng2021one}:} STAR uses a shared factorized network and a domain-specific factorized network in each domain to predict CTR for enumerated domains. In this paper, we use the time-period as the scenario division indicator, which is divided into breakfast, lunch, afternoon tea, dinner and night, 5 scenarios in total.
(5) \textbf{M2M\cite{zhang2022leaving}:} M2M designs multiple meta units to learn inter-scenario correlations and scenario-specific features to accomplish multi-task and multi-scenario advertiser modeling. In our experiments, we use spatiotemporal information as the input of meta unit to achieve dynamic modeling of spatiotemporal data.
(6) \textbf{APG\cite{APG}:} APG realizes self-wise adaptive modeling by building
parameter generation network and parameter adaptor to improve the performance of CTR prediction.

\begin{figure*}[tb]
  \centering
  {
      \includegraphics[width=1.9\columnwidth]{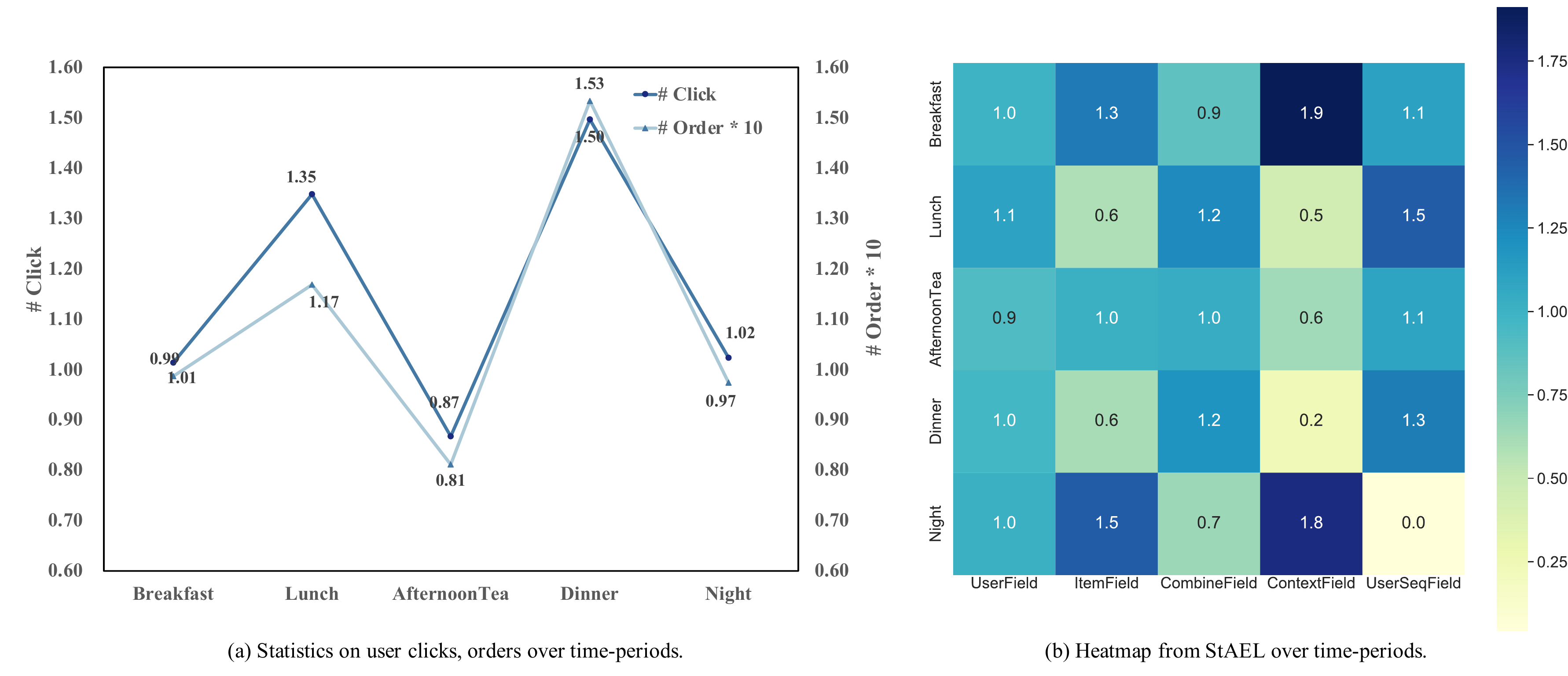}
  }
  \caption{Statistics and importance heatmap from StAEL over different time-periods. (a) Statistics on user clicks and orders over time-periods, demonstrating user activity. (b) The heatmap depicts the spatiotemporal weight $\alpha_j$ of each feature field over different time-periods. (Best viewed in color.)}
  \label{fig:tt_att}
\end{figure*}

\subsubsection{Evaluation Metrics}
In our experiment, we use Area Under Curve (AUC), Normalized Discounted Cumulative Gain(NDCG3, NDCG10) and Logloss, which have been  widely used in  recommendation systems. In addition, in order to further evaluate the fitting capability of the model on the spatiotemporal data, we proposed TAUC(Time-wise AUC)
 and CAUC(City-wise AUC) as our additional metrics. The computations of TAUC and CAUC are defined as follows:
\begin{itemize}
    \item TAUC: TAUC is the weighted average value of AUC for each time-period, it measures the quality of ranking in each time-period:
        \begin{align}
        {\mathbf{TAUC}}= \frac{\sum_{t=1}^{T}impression_t * AUC_t}{\sum_{t=1}^{T}impression_t}
        \end{align}
    where $impression_{t}$ is the number of exposures at $t^{th}$ time-period and $AUC_{t}$ is the AUC of $t^{th}$ time-period.
    \item CAUC: CAUC is the weighted average value of AUC for each city, it measures the quality of ranking in each city:
        \begin{align}
        {\mathbf{CAUC}}= \frac{\sum_{c=1}^{C}impression_c * AUC_c}{\sum_{c=1}^{C}impression_c}
        \end{align}
    where $impression_{c}$ is the number of exposures at $c^{th}$ city and $AUC_{c}$ is the AUC of $c^{th}$ city.
\end{itemize}

\subsubsection{Parameter Settings} Models in this paper are implemented with Tensorflow 1.4 in Python 2.7 environment. AdagradDecay\cite{duchi2011adaptive} is chosen as our optimizer for model training. The learning rate starts with 0.001 and increases over 1M steps to 0.012. The activation function of neural network is set to LeakyReLU and the batchsize is set to 1024. We use a warm-up\cite{he2016deep} technique for all models to prevent overfitting in the early training and maintain training stability. In addition, we averaged the results of all the studies after five repetitions.




\begin{figure*}[tb]
  \centering
  {
      \includegraphics[width=1.9\columnwidth]{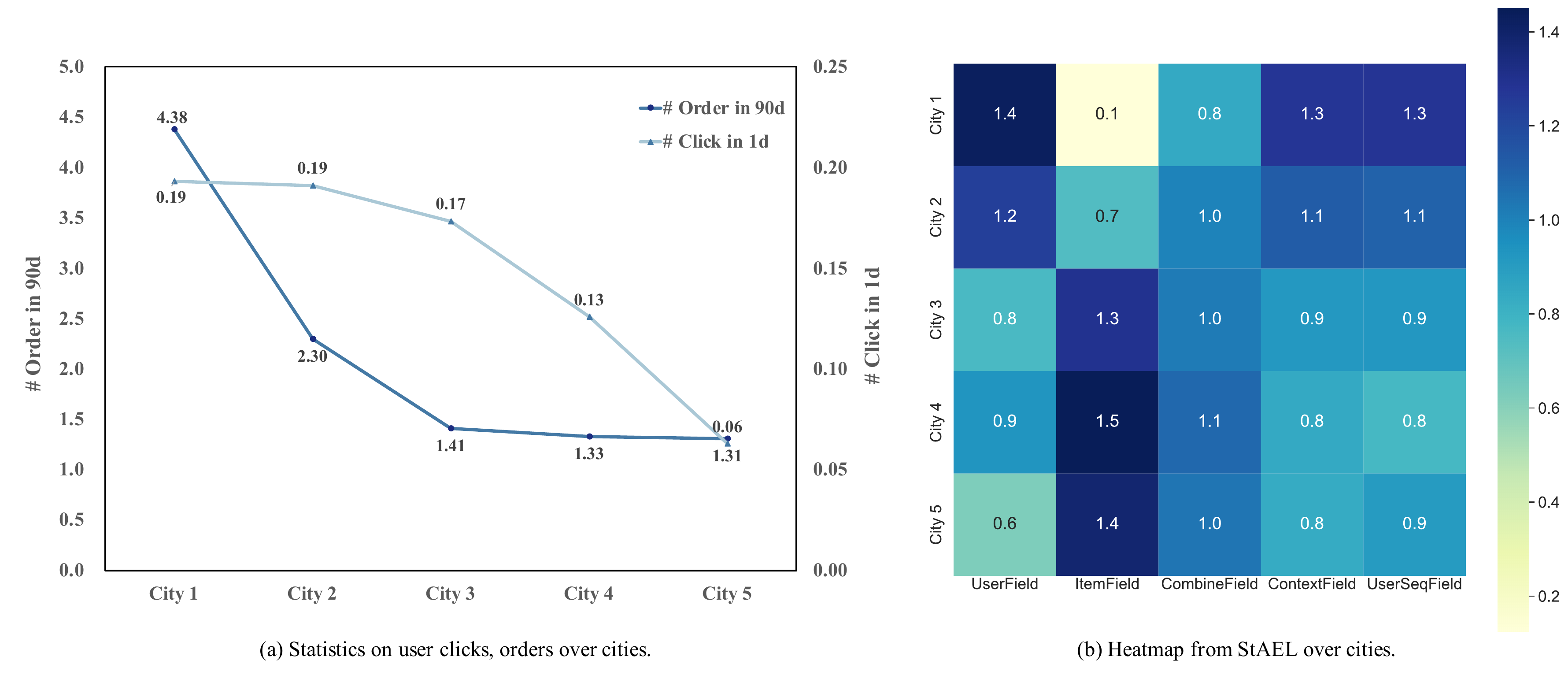}
  }
  \caption{Statistics and importance heatmap from StAEL over different cities. (a) Statistics on user clicks and orders over cities, demonstrating user activity. (b) The heatmap depicts the spatiotemporal weight $\alpha_j$ of each feature field over different cities. (Best viewed in color.)}
  \label{fig:location_att}
\end{figure*}

\begin{table}[tbp]
\centering
  \caption{ablation study on the impact of our three modules. StAEL, StSTL and StABT denote spatiotemporal-aware embedding layers, spatiotemporal semantic transformation layer and spatiotemporal adaptive bias tower, respectively.}\label{tab:ablation}
  \begin{tabular}{ccccc}
  \hline
    \multirow{2}*{Modules}&\multicolumn{4}{c}{Ele.me} \\ 
    \cline{2-5}
    ~&{AUC}&{TAUC}&{CAUC}&{Logloss}\\
  \hline
    {w/o StAEL}& 0.7363 & 0.7350 & 0.7340 & 0.1340 \\
    {w/o StSTL}& 0.7355 & 0.7341 & 0.7330 & 0.1547\\
    {w/o StABT}& 0.7349 & 0.7334 & 0.7323 & 0.1341\\
    \hline
    \textbf{BASM}&\textbf{0.7373}&\textbf{0.7360}&\textbf{0.7348}&\textbf{0.1339}\\
  \hline
  \end{tabular}
 \end{table}

\subsection{RQ1: Offline Performance Comparison}
 
Table \ref{tab:PerComp} indicates that BASM outperforms all State-Of-The-Art (SOTA) methods in all metrics on both industrial and public datasets.
In particular, among our two proposed metrics (TAUC and CAUC), our method achieves the state-of-the-art results in comparison with other methods, proving that BASM has better modeling effect on spatiotemporal data.
Specifically, on \emph{Spatiotemporal Public Data}, BASM achieves a further significant improvement\footnote{Note that the 0.1{\%} absolute AUC gain is already considerable in practical application \cite{WDL,autoint,zhou2018deep,APG}.} compared to the best comparison method M2M, $i.e.$, AUC +0.46\%, TAUC +0.39\%, and CAUC +0.25\%.

Compared with static parameter based methods (Wide\&Deep, DIN and AutoInt), dynamic parameter based methods (STAR, M2M, APG and ours) have a relatively significant improvement, because static parameter based methods simply mix all samples to train a set of model parameters, which makes it difficult to capture diversities in different spatiotemporal contexts.


However, dynamic parameter based methods either ignore the effects of spatiotemporal factors or rely heavily on predefined parameter spaces, limiting the ability of models to fit spatiotemporal data. For instance, our model has gained {0.42\%} and {0.61\%} AUC improvement compared to STAR on \emph{Ele.me} and \emph{Spatiotemporal Public Data}, respectively. STAR mainly models an exclusive network for each domain, but it is difficult to enumerate all domains for spatiotemporal scenarios. The artificial pre-defined scenario division limits the performance of STAR in spatiotemporal data. When compared to M2M and APG in \emph{Spatiotemporal Public Data}, our model improves by 0.46\% and 0.98\% in AUC and decreases by 0.0008 and 0.0037 in Logloss,  respectively. On the one hand, this is due to the fact that the dynamic parameters do not thoroughly investigate the influence of spatiotemporal factors, and their direct application will result in poor interpretability. On the other hand, M2M and APG do not take explicit debiasing solutions for distributions in various spatiotemporal scenarios into consideration.

It demonstrates that 
1) for characterizing differences in spatiotemporal distributions, it is crucial to strengthen spatiotemporal factors as dynamic parameters;
2) the spatiotemporal scenarios are difficult to enumerate, and the method of dynamic modeling with multiple sets of parameters is challenging;
3) in OFOS, there are natural differences in CTR under different time and location (spatiotemporal bias), which cannot be ignored in the modeling of spatiotemporal data. 
 
\subsection{RQ2: Ablation Study and Visualization}
To demonstrate the effectiveness of the BASM modules, ablation studies have been conducted on \emph{Ele.me}. The result is illustrated in Table \ref{tab:ablation}.


\subsubsection{Spatiotemporal-Aware Embedding Layer}
This module has outstanding performance in TAUC, CAUC and AUC, as shown in Table~\ref{tab:ablation},  the Spatiotemporal-Aware Embedding Layer improves TAUC by +{0.10\%}, CAUC by +{0.08\%}, AUC +{0.10\%}. It shows that the dynamic perception of different spatiotemporal contexts in the feature embedding layer of the model , which provides significant help for the subsequent task learning of the network.


\subsubsection{Spatiotemporal Semantic Transformation Layer}
We remove the Spatiotemporal Semantic Transformation Layer (w/o StSTL) from BASM to evaluate the performance. We observed that when it was removed, the auc fell by 0.18\% and the Logloss increased by {0.0208}. This shows that the addition of this module can further enhance the semantic representation of different spatiotemporal contexts. Nevertheless, we observe that the effect decrease of removing this module is much larger than the other two modules, which indicates that importance of semantic transformation on modeling spatiotemporal data.

\subsubsection{Spatiotemporal Adaptive Bias Tower}

In order to capture diverse spatiotemporal bias, we introduce Spatiotemporal Adaptive Bias Tower to reduce the difficulty of spatiotemporal distinction modeling. According to Table \ref{tab:ablation}, the removal of this module (w/o StABT) resulted in {0.24\%} reduction in AUC, 0.26\% in TAUC and 0.25\% in CAUC, validating the effectiveness of the Spatiotemporal Adaptive Bias Tower.



\subsubsection{Visualization about the spatiotemporal weight $\alpha_j$}
\begin{figure*}[tp]
\centering
\subfigure[Base model.]{\includegraphics[width=0.8\columnwidth]{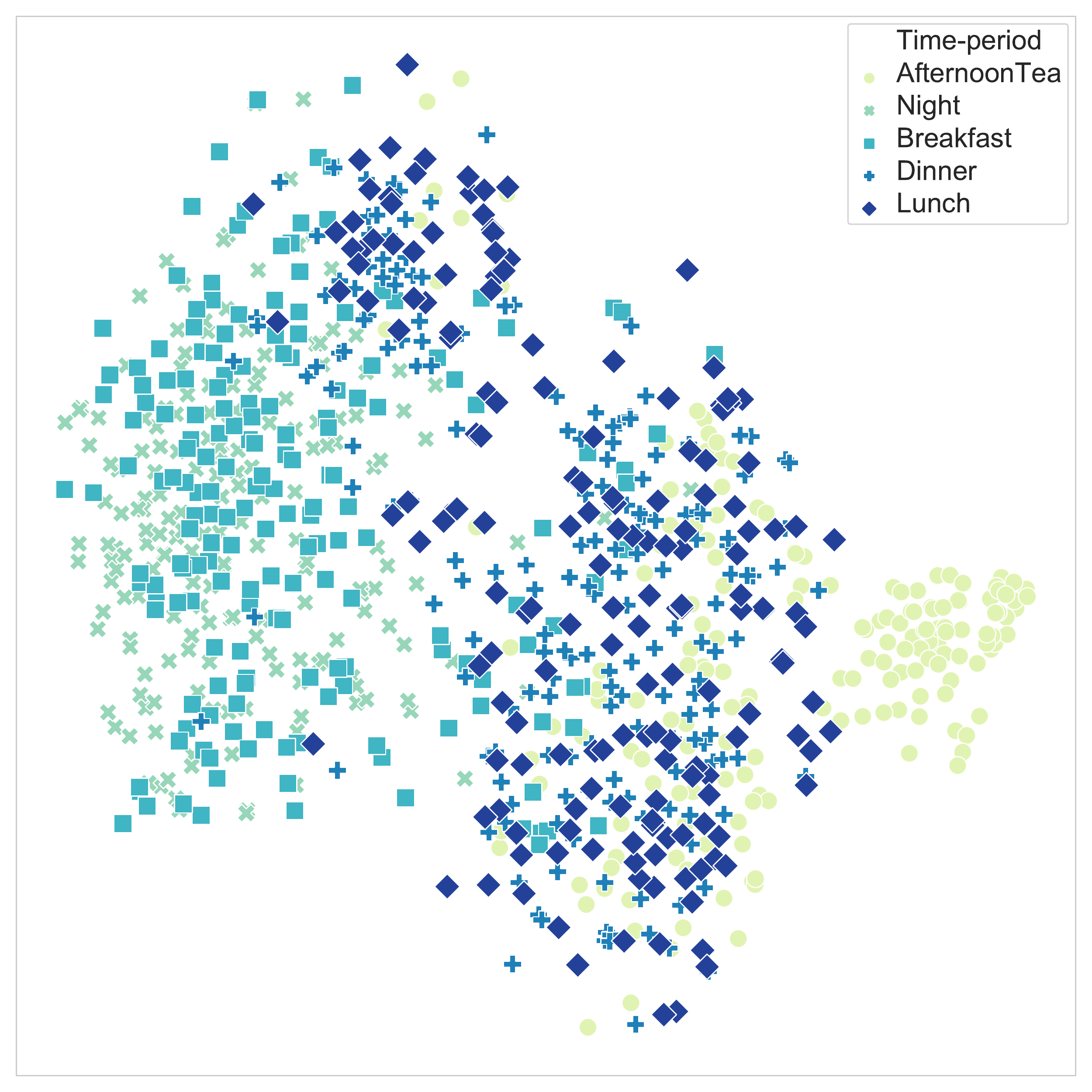}}
\subfigure[BASM.]{\includegraphics[width=0.8\columnwidth]{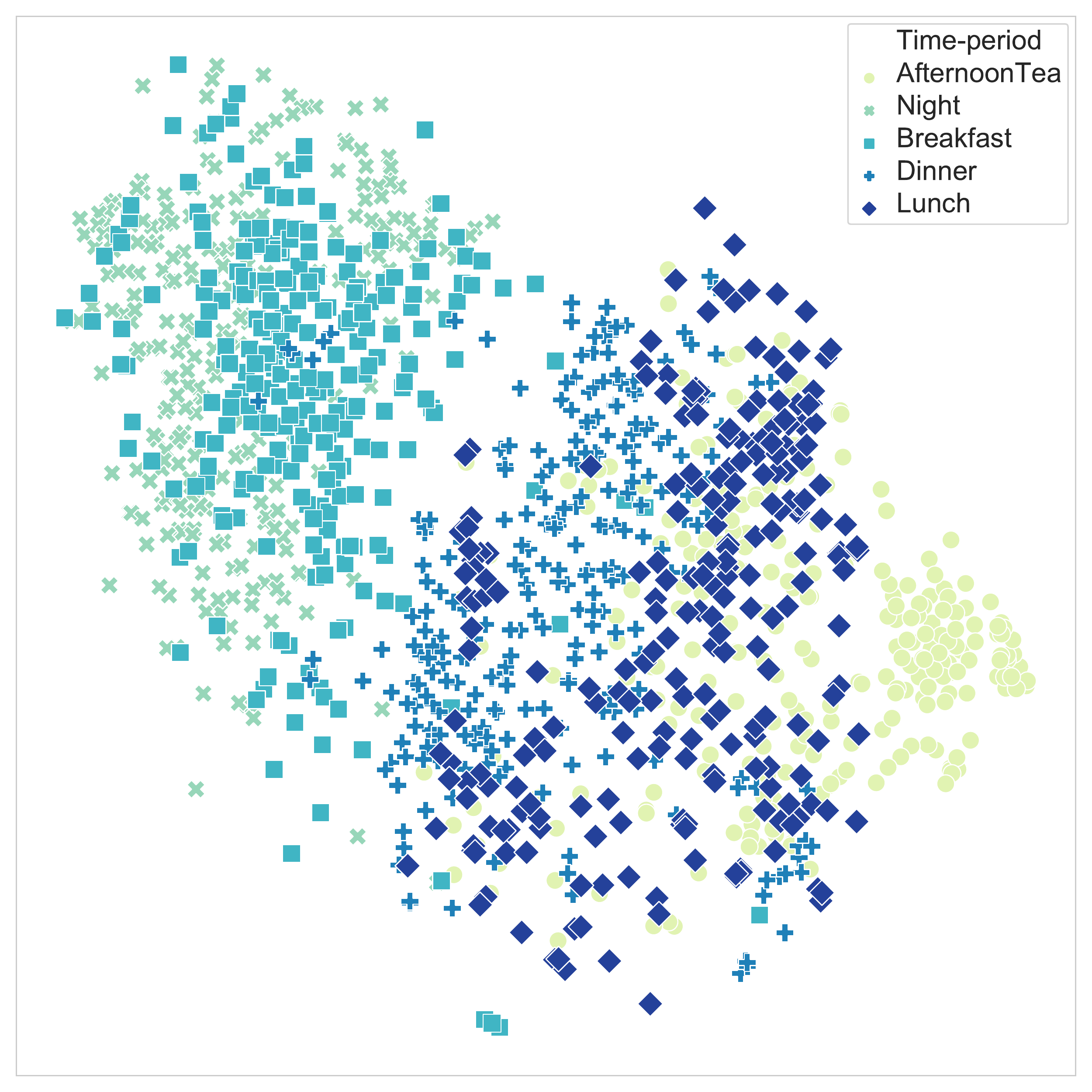}}
\caption{The t-SNE visualization of instance features generated by Base model (variation of DIN) and BASM over different time-periods.  (Best viewed in color.)}
\label{fig:vis_timetype}
\end{figure*}


In addition to proving the effectiveness of Spatiotemporal-Aware Embedding Layer (StAEL), we verify whether our gated attention could learn reasonable adaptive weights to perceive spatiotemporal context. We recorded the spatiotemporal weights $\alpha_j$ learned by StAEL in different spatiotemporal scenarios, and the visualized the heatmap of the spatiotemporal weights $\alpha_j$ in different time-preriods and cities, shown in Fig.~\ref{fig:tt_att} and \ref{fig:location_att}, respectively.

From Fig.~\ref{fig:tt_att}(a), we can see that the averages of user's clicks and orders is higher for lunch and dinner compared to others, indicating that users are more active at lunch and dinner. Fig.~\ref{fig:tt_att}(b) shows that, at lunch and dinner, our StAEL learns higher spatiotemporal weights $\alpha_j$ for the user field, user behavior sequence field and combine feature field. This confirms that StAEL assigns higher spatiotemporal weights to user-side features during user active periods, which is intuitive and consistent with data performance. The left image shows that users have weaker click and purchase intentions at breakfast and night, while the heatmap shows that StAEL increases the spatiotemporal weights  $\alpha_j$ of item feature field and context feature field, and the two trends are consistent. 

Moreover, we choose five typical cities to examine the impact from various locations. The number of users decreases from  City 1 to City 5. Observed from Fig.~\ref{fig:location_att}(a), it is obvious to find that the average value of orders within 90 days and the average value of clicks within 1 day from City 1 to City 5 decreased, which can be regarded as the difference in the above-mentioned city-wise user activity.
Correspondingly, as shown in Fig.~\ref{fig:location_att}(b), we can see that our method detects changes in the spatiotemporal weight $\alpha_j$ of feature field, $i.e.$, with the increase of user activity, the weight of the feature fields of user profile and user behavior sequences increases, while the weight of item feature fields decreases.

Based on the aforementioned investigation, our method is competent in learning the significance of feature fields for various data performance in different spatiotemporal scenarios, which also exhibits its superior capability in spatiotemporal contexts awareness.


\begin{figure*}[tp]
\centering
\subfigure[Base model.]{\includegraphics[width=0.8\columnwidth]{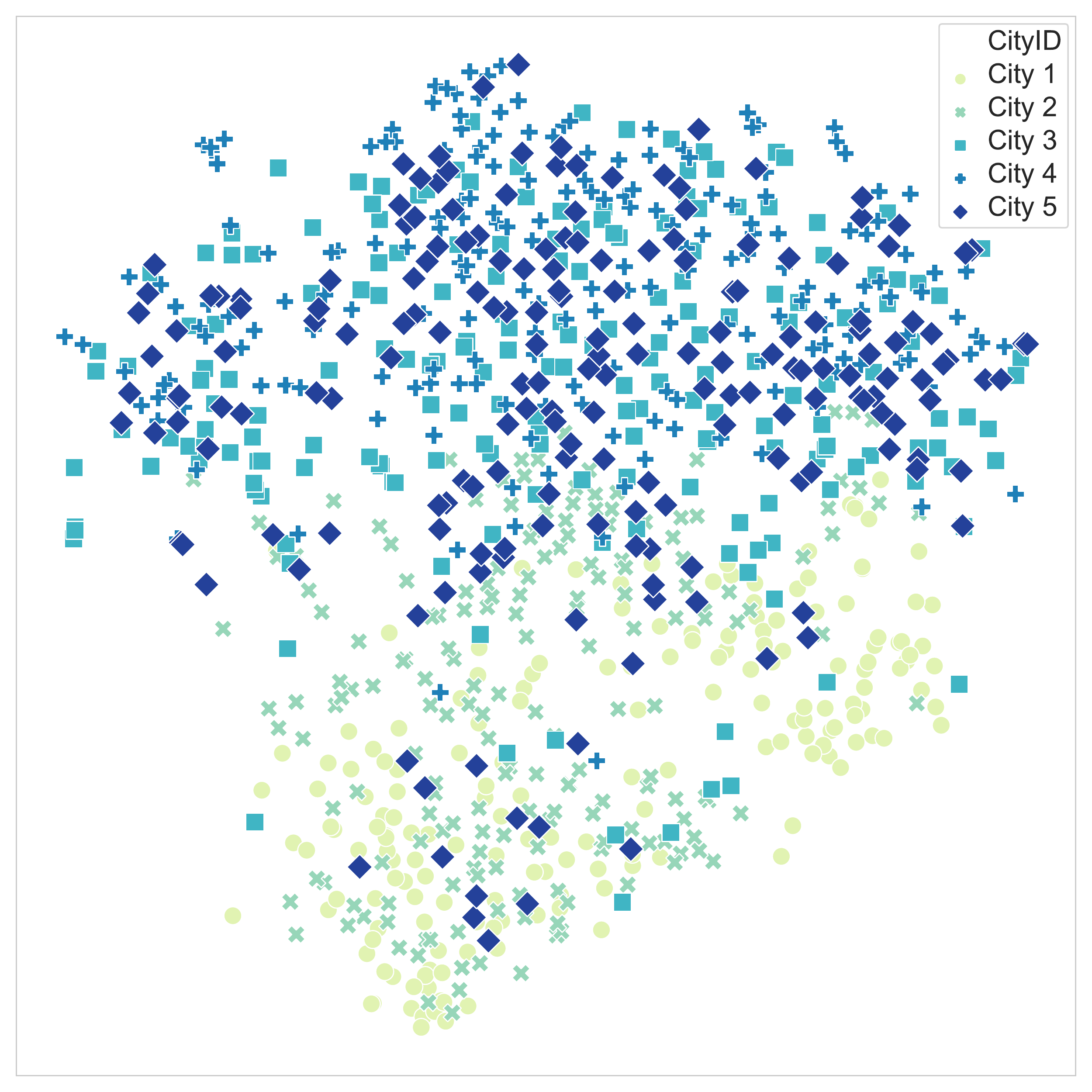}}
\subfigure[BASM.]{\includegraphics[width=0.8\columnwidth]{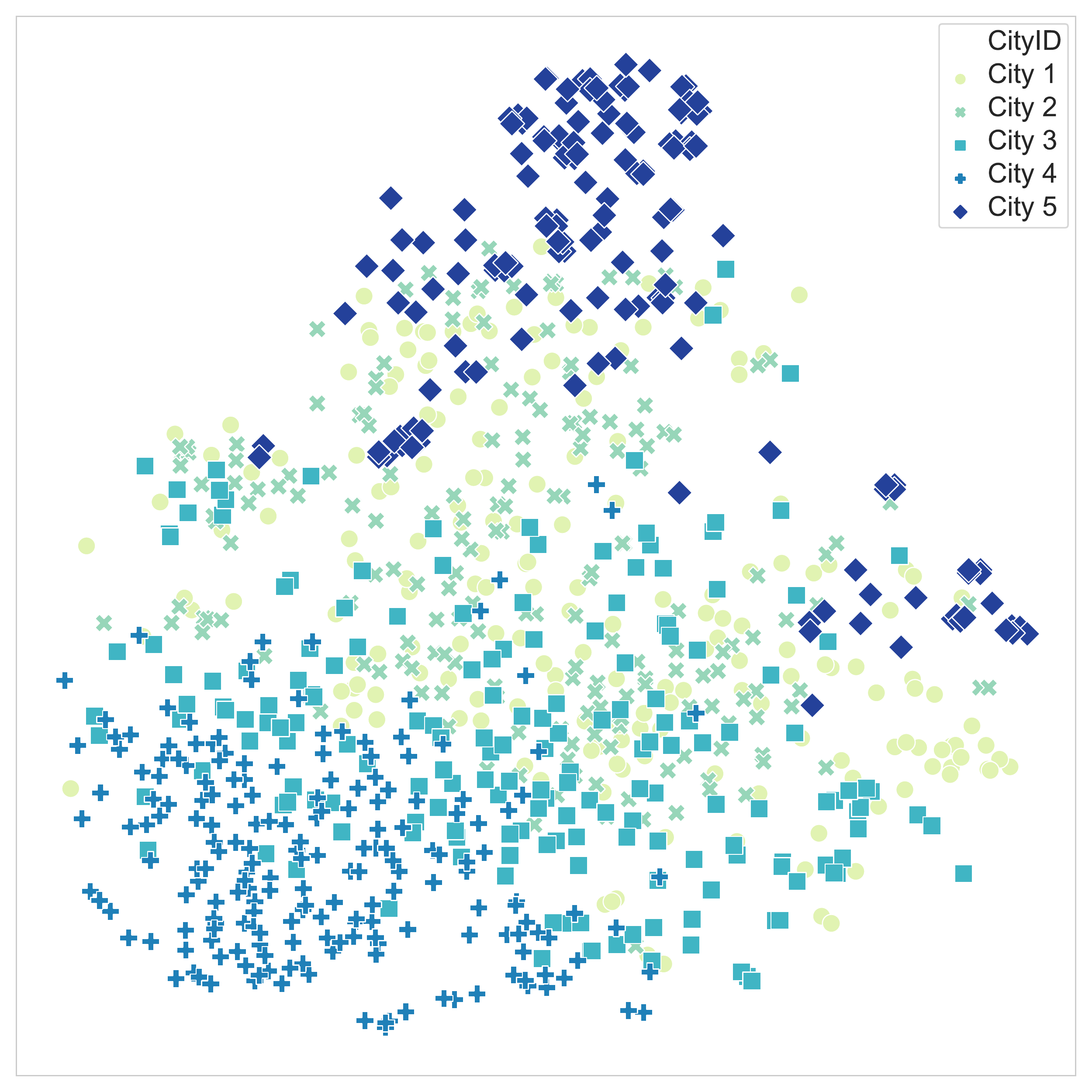}}
\caption{The t-SNE visualization of instance features generated by Base model (variation of DIN) and BASM over different cities. (Best viewed in color.)}
\label{fig:vis_location}
\end{figure*}

\begin{figure*}[tp]
\centering
\subfigure[Online exposure ratios and CTRs over time-periods.]{\includegraphics[width=0.9\columnwidth]{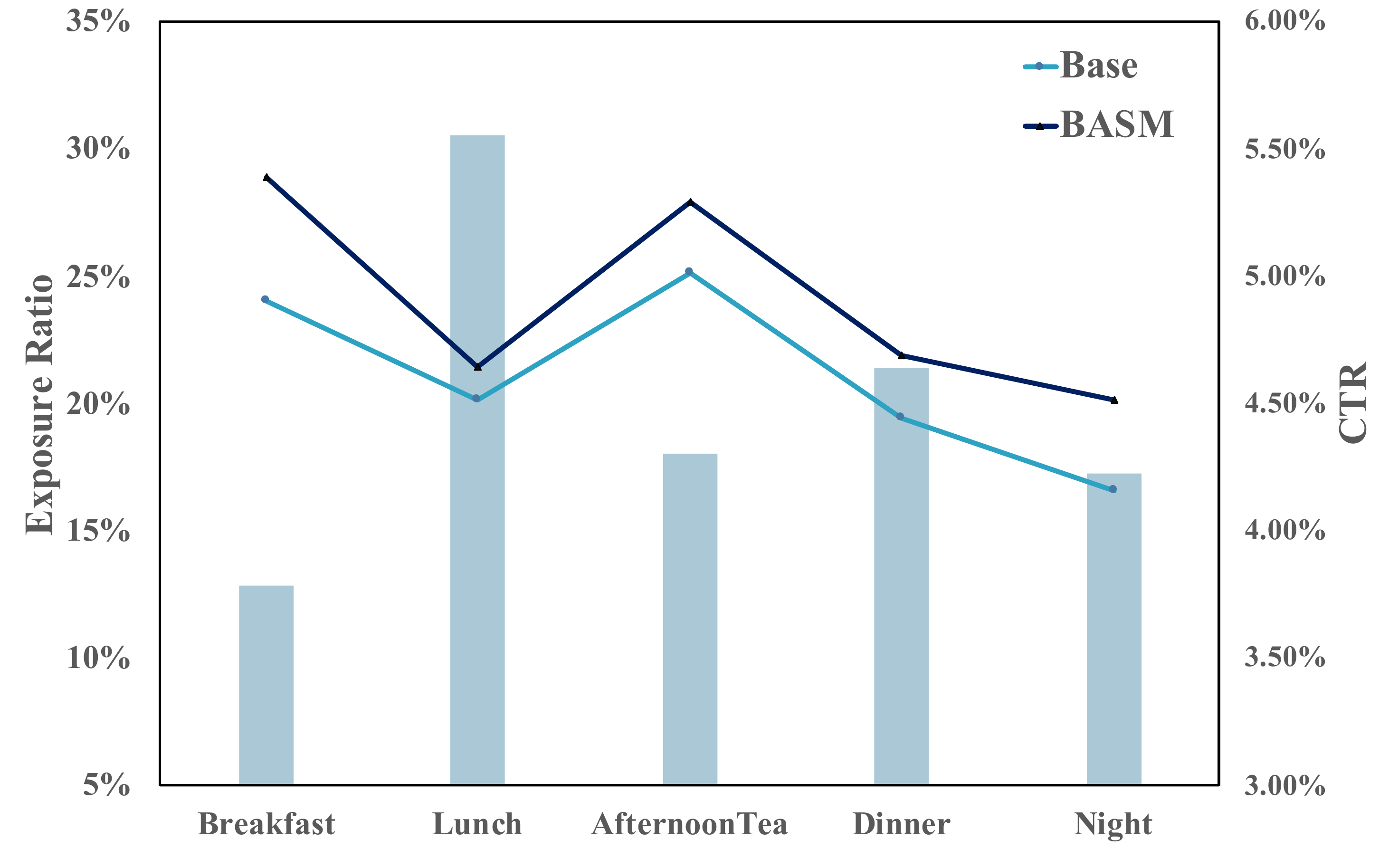}}
\subfigure[Online exposure ratios and CTRs over cities.]{\includegraphics[width=0.9\columnwidth]{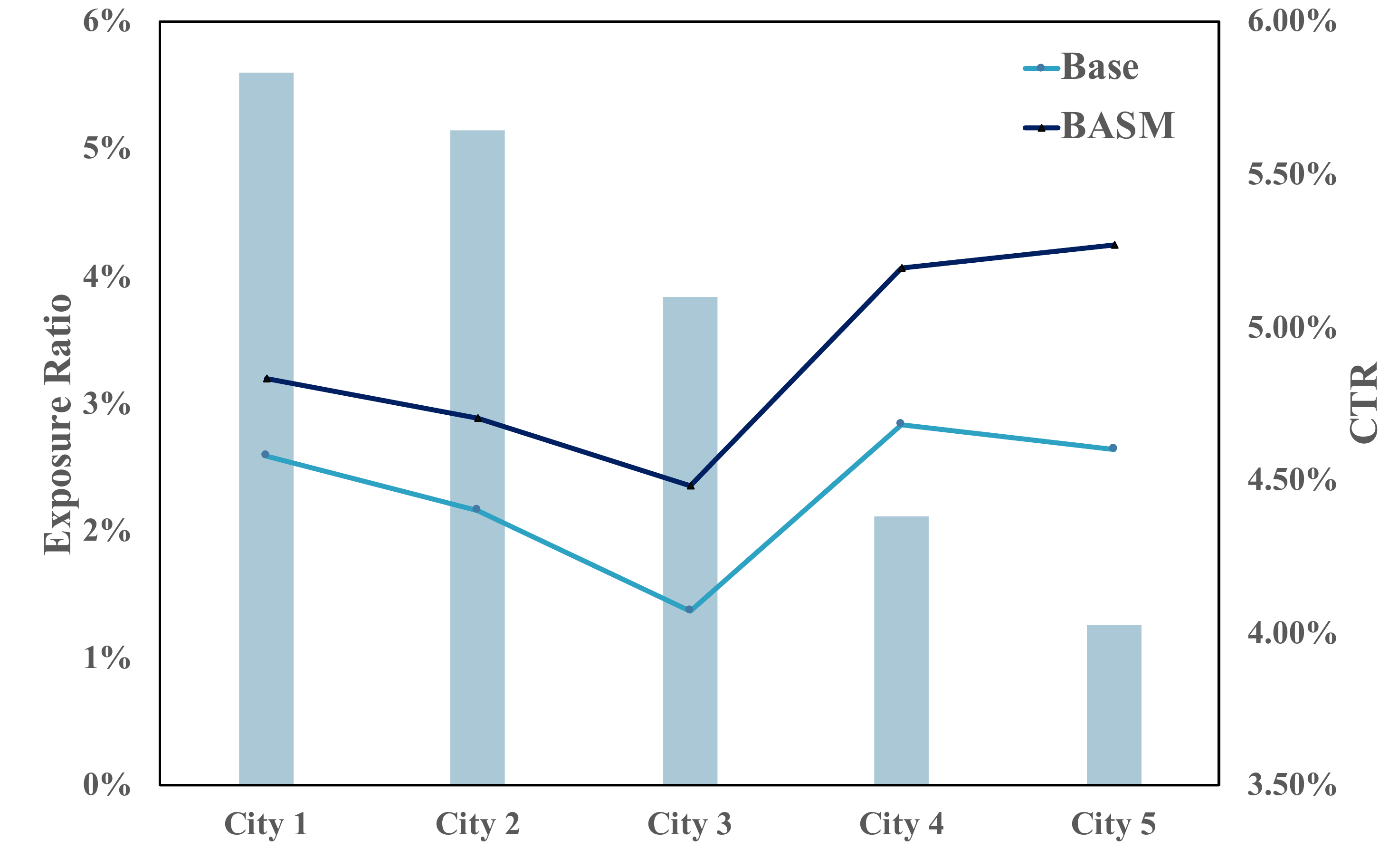}}
\caption{Online exposure ratios and CTRs of BASM and the base model (variation of DIN) over
time-periods or cities in one week in Aug 2022. The CTR improvement
is more significant in time-periods and cities with smaller exposure ratios.}
\label{fig:timecityab}
\end{figure*}

\subsubsection{Visualization about the model embedding} The t-SNE visualization of final embeddings between Base model (variation of DIN, detail can be seen in Section~\ref{sec:ab_test}) and BASM over different time-periods and different cities can be seen in Fig.~\ref{fig:vis_timetype} and  Fig.~\ref{fig:vis_location}, respectively. We can find that our BASM makes the instances in different time and location more convergent within the class and more dispersed among the classes.

Specially, compared with Fig.~\ref{fig:vis_timetype}(a), instances between dinner and lunch would are segmented by a clear manifold structure, as shown in Fig.~\ref{fig:vis_timetype}(b). Those samples with similar semantic are closely in the hyperspace, $i.e.$, samples of night and samples of afternoon tea. What's more, it is observed from Fig.~\ref{fig:vis_location}(b) that clusters of instances in different cities are more convergent, while mixed in Fig.~\ref{fig:vis_location}(a). Those prove that our BASM can better adapt to different spatiotemporal distributions.

\subsection{RQ3: System Efficiency}
In this section, we evaluate the time and memory efficiency of both BASM and its comparison methods by using the \emph{Ele.me} dataset. Experimental results are displayed in Table~\ref{tab:sys_eff}, and we observe that static parameter based methods ($i.e$, Wide\&Deep, DIN, and AutoInt) consistently get shorter training time and memory usage compared with dynamic parameter based methods, since these static parameter based methods simply mix all samples to train a set of model parameters, which is at the cost of reducing the performance of the model. 
However, in our proposed method, we reduce the overall parameter size of BASM through the rational model structure design and matrix decomposition method. Compared with other dynamic parameter based methods ($i.e$, STAR, M2M, and APG), BASM achieves the best model results with the lowest training time and memory consumption on the \emph{Ele.me} dataset.

\subsection{RQ4: Online A/B Test}
\label{sec:ab_test}

In August 2022, we conducted an online experiment by deploying BASM to the recommendation scenario on the homepage of Ele.me for one week.
We also deployed the base model, the variation of DIN, mainly consisting of three Multi-head Target Attention modules on the user's long/short/ realtime historical behavior sequence. Click-through rate, or CTR, was used to evaluate the performance of an online experiment, which was defined as the number of clicks over the number of item impressions. Strictly online A/B experiments are shown in Table~\ref{tab:ABDay}.  We can see that the proposed BASM consistently outperforms the base model. On average, our model improves CTR by \textbf{6.51\%} compared to the base model, which demonstrates the effectiveness of BASM in industrial spatiotemporal scenarios. BASM has been deployed on Ele.me, Alibaba's online food delivery platform, and currently serves more than 100 million users.

\begin{table}[tp]
\centering
  \caption{The training time per epoch and memory cost.}\label{tab:sys_eff}
  \begin{tabular}{ccc}
  \hline
    \multirow{2}*{Methods}&\multicolumn{2}{c}{Ele.me} \\ 
    \cline{2-3}
    ~&{Time / Epoch (min)}&{Memory (G)}\\
  \hline
    {Wide\&Deep}& 168 & 12.3 \\

    {DIN}& 183 & 14.2 \\

    {AutoInt}& 189 & 13.1 \\

    {STAR}& 301 & 35.3 \\

    {M2M}& 299 & 34.3 \\

    {APG}& 567 & 48.2 \\
     \hline
    \textbf{BASM}&\textbf{256}&\textbf{32.1}\\
  \hline
  \end{tabular}
 \end{table}


\subsection{RQ5: Online Spatiotemporal Result Analysis}
We analyze the performance of BASM and the base model at different time periods (breakfast, lunch, afternoon tea, dinner and night) and cities with different traffic scales. As illustrated in Fig.~\ref{fig:timecityab}, We found that: 1) BASM achieves
consistent improvement in all time-periods and cities; 2) the CTR improvement is more significant in time-periods and cities with smaller exposure ratios.
These results demonstrate the effectiveness of the Bottom-up Adaptive Spatiotemporal Model in fitting the variance of spatiotemporal data distribution and capturing the diversity of user interest under different location and time. 


\section{SYSTEM IMPLEMENTATION AND DEPLOYMENT}
This section introduces the system implementation and deployment of BASM on Ele.me, one of major online food recommendation platform of Alibaba. The system implementation and deployment mainly include offline training and online serving, as shown in Fig.~\ref{fig:deployment_of_BASM}.

\begin{table}[tp]
\centering
  \caption{Online A/B performances for consecutive 7 days in Aug 2022.}\label{tab:ABDay}
  \begin{tabular}{cccc}
  \hline
    \multirow{2}*{Day} &\multicolumn{2}{c}{CTR}&\multirow{2}*{Relative Improvement} \\ 
    \cline{2-3}
    ~&{Base model}&{BASM}\\
  \hline
    1&4.26 &4.63 & 8.69\%\\
    2&4.36 &4.70 &7.80\% \\
    3&4.35 &4.56 &4.83\%\\
    4&4.57 &4.86 &6.35\% \\
    5&4.98 &5.24 &5.22\%\\
    6&4.69 &5.03 &7.25\%\\
    7&5.09 &5.35 &5.11\%\\
    \hline
    Avg&4.61 &4.91 &6.51\%\\
  \hline
  \end{tabular}
 \end{table}

\subsection{Offline Training}
We first store the collected online behavior of users on the self-developed MaxCompute platform (MCP) in log format. By processing the data, train and test datasets can be generated . Through distributed training on Alibaba Online Learning Platform (AOP) and performance evaluation, we finally deploy BASM on a Real-Time Prediction (RTP) platform for online serving.

\subsection{Online Serving}
\begin{figure}[tp]
\centerline{\includegraphics[width=1.\columnwidth]{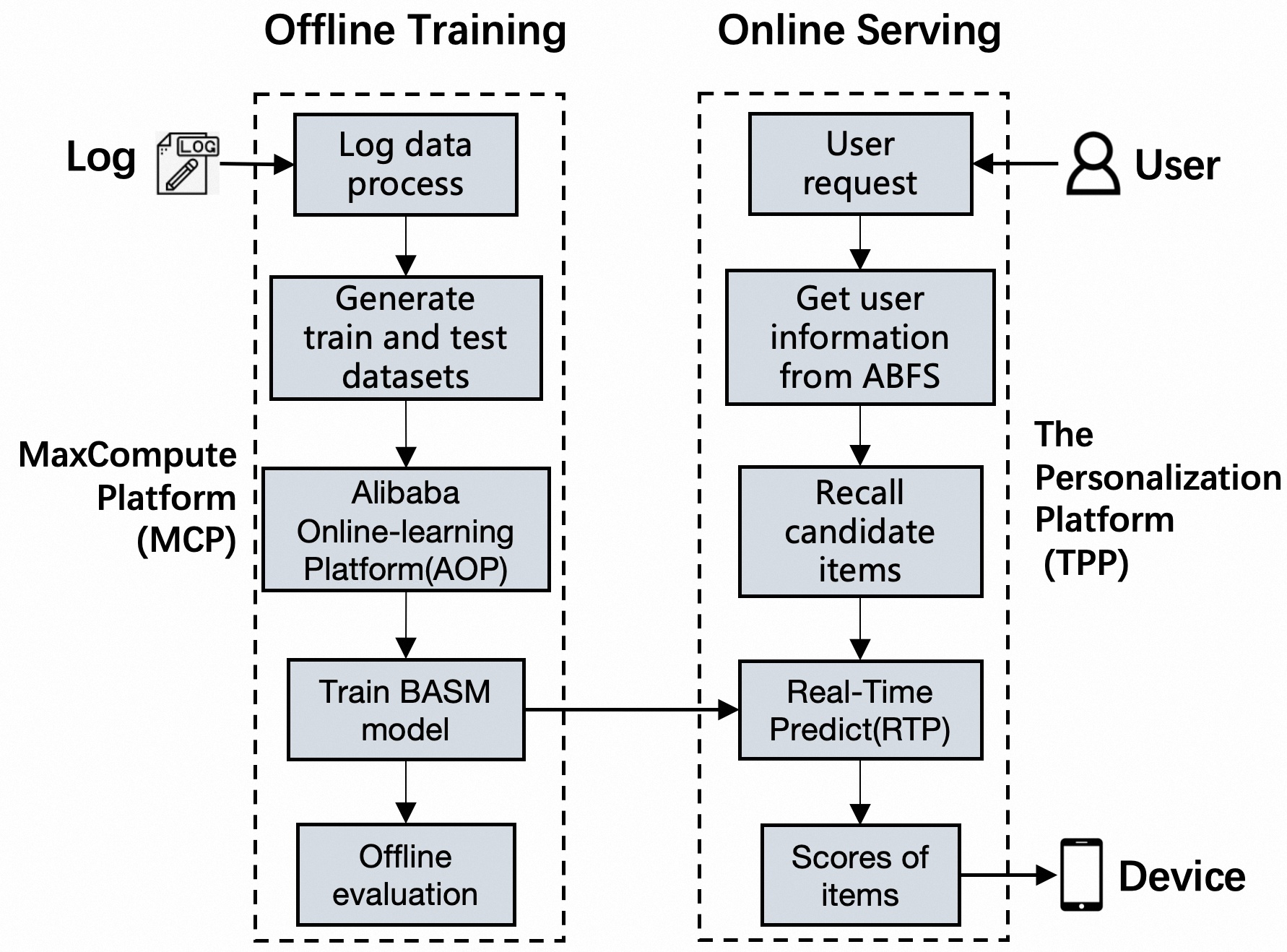}}
\caption{Offline Training and Online Deployment of BASM}
\label{fig:deployment_of_BASM}
\end{figure}

As shown in Fig.~\ref{fig:deployment_of_BASM}, the online serving  of BASM is implemented on the Personalization Platform (TPP). When a user logs in to the Ele.me APP, TPP obtains user-side features, including user basic features and behavior sequences, by calling Alibaba Basic Feature Server (ABFS). The candidate items are recalled based on Location-based Service, then  user features, context features and candidate items are fed to RTP for scoring prediction. Finally, according to the RTP score, the top k items are returned for exposure.

\section{Related Works}
\subsection{Feature Representation Learning}


Many methods of feature representation have achieved great benefits. Wide\&Deep (WDL) \cite{WDL} jointly trains wide linear part and MLP to combine the benefits of memorization and generalization. DeepFM \cite{DFM} replaces the wide part of WDL with factorization machine to alleviate manual efforts in feature engineering. AutoInt \cite{autoint} applies stacked multi-head self-attention layers to model high-order feature interactions. However, these methods learn a fixed representation of each feature regardless of the different importance of each feature under different contexts. 


\subsection{Dynamic Network}
The LHUC \cite{Lhuc} method scales the activation of the hidden layer by treating each speaker as an element-wise multiplier, so that each speaker learns a specific hidden unit contribution to improve the speech recognition performance. CAN \cite{CAN} achieves explicitly learning co-action representation  by extending the user feature representation by using a Multi-layer perceptron whose parameters are dynamically generated from items. APG \cite{APG} designs a new learning paradigm to capture custom and common patterns by dynamically generating parameters across different instances.


\subsection{Spatiotemporal Systems}
Most current spatiotemporal systems are primarily aimed at the next point-of-interest recommendation. STAN \cite{STAN} proposed a spatiotemporal bi-attention model to activate the time interval and distance between the GPS location to learn the regularities between non-adjacent locations and non-contiguous visits. ST-PIL \cite{ST-PIL} proposed two levels of attention to fully consider the spatial-temporal context for periodic interest learning. In addition, user behaviors naturally depend on the spatiotemporal information in location-based service platforms, such as Ele.me and Meituan. StEN \cite{StEN} proposed three modules to extract spatiotemporal preference, where the Spatiotemporal-aware Target Attention mechanism employed different spatiotemporal information to generate different parameters and feed them into target attention to improve the personalized spatiotemporal awareness of the model. And TRISAN \cite{TRISAN} captures the geographic information and time information through an attention-based fusion mechanism to realize the CTR prediction in location-based search.

\subsection{Debias}
Bias is common in recommendation because user behavior is based on observational exposure, many research effort on debiases. PAL \cite{guo2019pal} proposes a Position-bias Aware Learning framework for CTR prediction in a live recommender system. DICE \cite{zheng2021disentangling} proposes a general framework to disentangle interest and conformity(popularity bias), to capture the variety of conformity, which is independent with user interest, has a high robustness and interpretability.


\section{Conclusion}
In this paper, we propose a Bottom-up Adaptive Spatiotemporal Model(BASM) to adaptively fit the spatiotemporal data distribution, which further improves the fitting ability of the model. To address the challenge of spatiotemporal data distribution, we realize adaptive modeling of spatiotemporal data from the bottom embedding layer, middle semantic layer and top classification tower. Specifically, a spatiotemporal-aware embedding layer performs weight adaptation on field granularity in feature embedding, to achieve the purpose of dynamically perceiving spatiotemporal contexts. Besides, we propose a spatiotemporal semantic transformation layer to explicitly convert the concatenated input of the raw semantic to spatiotemporal semantic, which can further enhance the semantic representation under different spatiotemporal contexts. What's more, we introduce a novel spatiotemporal adaptive bias tower to capture diverse spatiotemporal bias, reducing the difficulty  to model spatiotemporal distinction. Extensive offline experiments and online A/B demonstrate effectiveness and efficiency of BASM. 

Although BASM is mainly modeled for spatiotemporal scenarios, we believe that it can be generalized to other scenarios with multiple data distributions in the future.

\bibliographystyle{IEEEtran}

\bibliography{./bib/paper}

\end{document}